\documentclass[runningheads]{llncs}

\usepackage{eccv}

\usepackage{eccvabbrv}

\usepackage{graphicx}
\usepackage{booktabs}
\usepackage{algorithm}
\usepackage{booktabs}
\usepackage{siunitx}
\usepackage{adjustbox}
\usepackage{subcaption}
\usepackage{makecell}
\usepackage{multirow}
\usepackage{lipsum}
\usepackage{wrapfig}
\usepackage{capt-of}
\usepackage{subcaption}
\usepackage{siunitx}
\usepackage{algorithmicx}

\usepackage{algpseudocode}

\usepackage{amsmath,amssymb,amsthm,bm}

\newcommand{\tr}{\mathrm{tr}}

\newcommand{\dd}{\mathrm{d}}

\theoremstyle{plain}

\theoremstyle{remark}
\usepackage[table,xcdraw]{xcolor}
\usepackage[accsupp]{axessibility}  

\usepackage{hyperref}

\usepackage{orcidlink}
\newcommand{\paratitle}[1]{\ifdefined\toggleParatitle\else\textcolor{blue}{#1}\fi}

\begin{document}

\title{Training-Free Refinement of Flow Matching with Divergence-based Sampling} 

\titlerunning{Flow Divergence Sampler}

\author{
Yeonwoo Cha~~~
Jaehoon Yoo~~~
Semin Kim   \\
~~Yunseo Park~~~   
Jinhyeon Kwon~~
Seunghoon Hong$^\dagger$
}

\authorrunning{Y.~Cha et al.}

\institute{
KAIST, South Korea\\
\medskip
\url{https://yeonwoo378.github.io/official_fds}
}

\maketitle
\begingroup
\renewcommand{\thefootnote}{\fnsymbol{footnote}}
\footnotetext{$\dagger$ Corresponding author.}
\endgroup

\vspace{-11pt}
\begin{abstract}

Flow-based models learn a target distribution by modeling a marginal velocity field, defined as the average of sample-wise velocities connecting each sample from a simple prior to the target data.
However, when sample-wise velocities conflict at the same intermediate state, this averaged velocity can misguide samples toward low-density regions, degrading generation quality. To address this issue, we propose the Flow Divergence Sampler (FDS), a training-free framework that refines intermediate states before each solver step.
Our key finding reveals that the severity of this misguidance is quantified by the divergence of the marginal velocity field that is readily computable during inference with a well-optimized model.
FDS exploits this signal to steer states toward less ambiguous regions. 
As a plug-and-play framework compatible with standard solvers and off-the-shelf flow backbones, FDS consistently improves fidelity across various generation tasks including text-to-image synthesis, and inverse problems.

\vspace{-5pt}

\end{abstract}

\vspace{-0.8cm}
\section{Introduction}
\vspace{-4pt}
Flow Matching (FM)~\cite{song2022ddim, lipman2023fm, esser2024sd3, peebles2023dit} have emerged as a powerful paradigm for modeling complex target distributions, enabling high-fidelity synthesis across image~\cite{rombach2022ldm, podell2023sdxl, xie2024sana}, audio~\cite{huang2023makeanaudio, liu2024audioldm2}, and video~\cite{singer2022makeavideo, ho2022imagenvideo}. 
Given a simple prior and a complex target data distribution, FM learns a mapping between them through a time-dependent velocity field.
Although each paired source-target sample induces its own sample-wise conditional velocities, FM learns a single marginal velocity at each intermediate state, which is an averaged velocity over all sample-wise velocities passing through that state.

While the marginal velocity field induces a valid mapping from source to target distribution, it introduces a fundamental limitation arising from directional conflicts among the underlying sample-wise velocities. 
Specifically, these velocities are locally multi-modal and can conflict, pointing in different directions at the same location, as illustrated in Fig.~\ref{fig:crossing}.
At such states, a single marginal velocity cannot align with any valid mode, but instead follows their average direction.
As a consequence, generative trajectories that pass through those states can drift toward low-density regions, producing blurry or degraded outputs.

Prior works~\cite{zhang2025hrm, guo2025vrm} have addressed this problem by redesigning the formulation to model the multi-modal distribution of these conflicting velocities.
While effective, these approaches require costly training, preventing the direct application of the strong off-the-shelf flow models that are already available.

In this paper, we ask a different question: can this problem be mitigated at inference time \emph{without} retraining the model?
Our core idea is to correct the trajectory at inference time rather than modifying the velocity field itself.
Before taking the next solver step, we update the current intermediate state to a nearby point where the same pre-trained vector field is expected to provide less ambiguous velocity, and then continue integration from there.
Since this refinement operates on the state rather than the model itself, it can be applied in a plug-and-play fashion to any off-the-shelf model and combined with standard solvers such as Euler and Heun.

The central challenge is that this local ambiguity is not directly observable at inference time, as evaluating the underlying dispersion of sample-wise velocities requires access to the training dataset.
Our key theoretical finding establishes that this dispersion admits a closed-form expression in terms of the divergence of the marginal velocity field.
In practice, this motivates a data-free surrogate based on the divergence of the pre-trained model. 
Building on this observation, we propose the Flow Divergence Sampler (FDS), which steers trajectories away from locally ambiguous regions during sampling.
Our extensive experiments show that, under matched compute budgets, FDS outperforms simply taking more solver steps in various settings.

Our contributions are summarized as follows: 
{\bf (1)} We show that the conditional mean-squared discrepancy between the marginal velocity and sample-wise velocities can be characterized through the divergence of the marginal velocity field, which in turn motivates a practical data-free proxy at inference time.
{\bf (2)} We introduce FDS, a plug-and-play sampling method that refines intermediate states toward locally less ambiguous regions without retraining of the underlying model. 
{\bf (3)} We demonstrate consistent improvements across diverse flow backbones, solvers, and various settings, including class-conditional generation, text-to-image synthesis, and inverse problems.

\section{Preliminaries}
\label{sec:preliminaries}
\vspace{-2pt}
\subsection{Flow Matching}

Flow Matching (FM)~\cite{lipman2023fm} constructs a continuous probability path $p_t$ that transports a simple prior distribution  $p_0 = \mathcal{N}(0, I)$ to a target data distribution $p_1$.
In general, this path can be defined through an interpolant between the noise sample $x_0 \sim p_0$ and a data sample $x_1 \sim p_1$ by

\begin{equation}
    x_t = \alpha_t x_1 + \beta_t x_0,~~x_t\sim p_t,
    \label{eq:interpolant}
\end{equation}
where $\alpha_t$ and $\beta_t$ are differentiable schedules that monotonically increase and decrease, respectively, satisfying $(\alpha_0, \beta_0) = (0, 1)$ and $(\alpha_1, \beta_1) = (1, 0)$ such that $x_t$ transitions from noise to data.

A velocity field that transports $p_t$ along this path can be characterized through the \emph{sample-wise velocity} of each interpolant:
\begin{equation}
    v_t(x_0, x_1) := \frac{dx_t}{dt} = \dot{\alpha}_t x_1 + \dot{\beta}_t x_0.
    \label{eq:instantaneous_velocity}
\end{equation}
%
Importantly, $v_t$ is a variable that varies across sample pairs $(x_0, x_1)$, not a vector field defined over $x_t$.

The Conditional Flow Matching (CFM) framework trains a neural network $u_\theta(x_t, t)$ to predict this velocity by minimizing
\begin{equation}
    \mathcal{L}_{\text{CFM}}(\theta) = \mathbb{E}_{t, x_0, x_1} \left[ \left\| u_\theta(x_t, t) - v_t \right\|^2 \right],
    \label{eq:cfm_objective}
\end{equation}
with $t \sim \mathcal{U}[0,1]$, $x_0 \sim p_0$, and $x_1 \sim p_1$.
Because \cref{eq:cfm_objective} is a least-squares objective, its pointwise minimizer at each $(x_t, t)$ is the conditional expectation
\begin{equation}
    u_t(x_t) := \mathbb{E}\left[ v_t \mid x_t \right],
    \label{eq:marginal_velocity}
\end{equation}
termed the \emph{marginal velocity field}.
Therefore, a well-trained flow model satisfies $u_\theta(x_t, t) \approx u_t(x_t)$.
At inference time, samples are generated by integrating the learned field
\begin{equation}
    \frac{dx_t}{dt} = u_\theta(x_t, t), \quad x_0 \sim p_0,
    \label{eq:inference_ode}
\end{equation}
from $t=0$ to $t=1$ over a finite number of discrete timesteps using standard numerical solvers, such as first-order Euler method or second-order Heun's method.

\vspace{-3pt}
\subsection{Discrepancy Between Marginal and Sample-wise Velocities}
\vspace{-1pt}
\label{sec:crossings}

\begin{figure}[!t]
  \centering
    \includegraphics[width=0.77\linewidth]{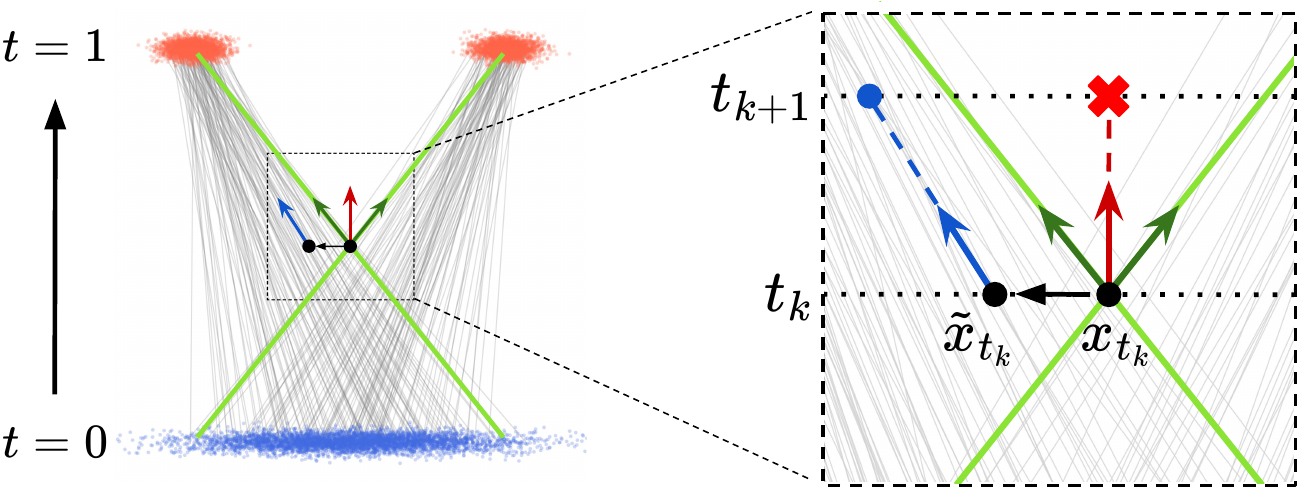}

  \vspace{-3pt}

  \caption{\textbf{Overview of FDS.} 
  In flow matching, multiple sample-wise velocities (green arrows) may pass through the same intermediate state $x_{t_k}$. 
 In such cases, the marginal velocity field becomes their average (red arrow), causing subsequent ODE integration to drift toward a low-density region. 
  To counteract this, FDS first updates the current state to a nearby state $\tilde{x}_{t_k}$ where the local directional conflict is weaker. 
  ODE integration from $\tilde{x}_{t_k}$ is then follows more reliable direction towards a mode (blue circle).  
  }
  \vspace{-11pt}
  \label{fig:crossing}
\end{figure}

In flow matching, interpolants from different sample pairings can pass through the same intermediate location $x_t$. 
At such crossings, the sample-wise velocities $v_t$ associated with each sample pair may point toward distinct target modes, while the marginal field $u_t(x_t)$ can only represent their average (\cref{eq:marginal_velocity}).
%
In that case, the marginal velocity may point toward a low-density region between modes rather than toward any valid mode (red cross in Fig.~\ref{fig:crossing}), causing the ODE trajectory during inference to produce blurry or degraded samples~\cite{zhang2025hrm, guo2025vrm}.

We quantify the severity of this degradation-inducing term through the residual of the optimal CFM predictor:
\begin{equation}
    \mathcal{L}^*_{\text{CFM}}(x_t, t) = \mathbb{E}\left[ \left\| u_t(x_t) - v_t \right\|^2 \;\middle|\; x_t \right],
    \label{eq:optimal_objective}
\end{equation}
which measures the \emph{discrepancy} between $v_t$ around its conditional mean $u_t(x_t)$.
The discrepancy is small when the sample-wise velocities are locally consistent, and large when multiple incompatible velocities coexist at the same state. 
Importantly, this residual is irreducible: it persists even for a perfectly optimized flow model using \cref{eq:cfm_objective}, since it reflects the intrinsic multi-modality of $v_t$ at $x_t$ rather than optimization error.

To address this issue, recent works~\cite{park2024caf, guo2025vrm, zhang2025hrm} redesign training to explicitly resolve crossings, by modeling the full multi-modal velocity distribution with an auxiliary network.
However, these approaches require costly training, often together with substantial architectural or objective modifications, and thus cannot directly leverage off-the-shelf flow models pre-trained on large-scale data.

\vspace{-2.5pt}
\section{Method}
\vspace{-3pt}
To address the issues discussed in Sec.~\ref{sec:crossings}, we seek an inference-time method that can be attached plug-and-play to off-the-shelf flow models. 
Rather than retraining the network or attempting to alter the learned marginal velocity field itself, we keep the pre-trained flow model fixed. 
Instead, our approach intervenes exclusively during the sampling phase, performing targeted spatial refinement at the exact same timestep $t$.


The central idea of our method is to refine the trajectory, rather than the velocity field. 
When interpolant crossings make the learned marginal velocity locally unreliable, we do not attempt to re-estimate the underlying multi-modal velocity distribution. 
Instead, at a solver state $x_t$, we seek a nearby state $\tilde{x}_t$ where the same pre-trained velocity field is expected to be more reliable, and then continue integration from $\tilde{x}_t$. 
In this sense, our method performs a spatial refinement of the intermediate state $x_t$, rather than a modification of the model itself (Fig.~\ref{fig:crossing}).

A natural objective for this refinement is the discrepancy $\mathcal{L}^*_{\text{CFM}}(x_t, t)$ (\cref{eq:optimal_objective}) between the marginal and sample-wise velocities, since it directly measures the local ambiguity of the marginal velocity at the current state.
However, this quantity is not observable at inference time, as evaluating \cref{eq:optimal_objective} requires access to the ground-truth sample-wise velocities $v_t$ that depend on training data.

To address this challenge, we introduce the \textbf{Flow Divergence Sampler (FDS)}. 
In this section, we first theoretically prove that this intractable, data-dependent discrepancy can be explicitly measured using a data-free surrogate at inference time (Sec.~\ref{sec:surrogate}). 
Building upon this, we outline our sampling strategy designed to seamlessly navigate the intermediate state $x_t$ toward reliable, low-discrepancy regions without excessive computational overhead (Sec.~\ref{sec:method_fds}).

\subsection{Flow Divergence as Discrepancy Surrogate}
\label{sec:surrogate}

Recall that the marginal velocity is the conditional mean (\cref{eq:marginal_velocity}). 
Thus, the discrepancy in \cref{eq:optimal_objective} measures the local spread of sample-wise velocities around their average $u_t(x_t)$. 
When multiple paths passing through the same $x_t$, this spread becomes large, and the averaged direction may become unreliable. 
Since the individual velocities are unobservable at inference-time, we instead examine how their average changes under small spatial perturbations of $x_t$. 
This leads to the spatial derivative of the marginal field, $J_{x_t}u_t(x_t)$, whose trace is the divergence $\nabla_{x_t}\!\cdot u_t(x_t)$. 

The following theorem shows that the ambiguity is exactly captured by the divergence of the marginal velocity, up to known schedule-dependent coefficients.

\begin{theorem}
For any $t$ such that $\alpha_t\neq 0$, the optimal CFM residual satisfies
\begin{equation}
\mathcal{L}^*_{\text{CFM}}(x_t, t)
= \mathbb{E} \left[ \left\| u_t(x_t)  - v_t\right\|^2 \;\middle|\; x_t \right]
=  \frac{\dot{\alpha}_t\beta_t - \alpha_t \dot{\beta}_t}{\alpha_t}\Big( \beta_t\nabla_{x_t}\cdot u_t(x_t) - \dot{\beta}_t d \Big),
\label{eq:main_div}
\end{equation}
where $d$ is the dimensionality of the data. 
\label{theorem:main}
\end{theorem}

The proof is provided in the App.~\ref{app:proof}. 
Thm. \ref{theorem:main} shows that the inaccessible discrepancy can be expressed entirely in terms of the divergence of the marginal velocity field and known schedule-dependent coefficients $\alpha_t$ and $\beta_t$.
Since all coefficients are constant with respect to $x_t$ for a fixed $t$, it implies that minimizing $\mathcal{L}^*_{\text{CFM}}(x_t, t)$ over nearby candidates is equivalent to minimizing the spatial divergence $\nabla_{x_t}\cdot u_t(x_t)$.

In practice, the true marginal velocity field $u_t$ is unavailable, so we replace it with the pre-trained model $u_\theta$, i.e., $u_\theta(x_t, t) \approx u_t(x_t)$, and define the following inference-time surrogate:

\begin{equation}
 \hat{\delta}_t(x) = \nabla_{x}\cdot u_\theta(x,t).
 \label{eqn:proxy}
\end{equation}
We use $\hat{\delta}_t(x)$ as a surrogate for local reliability: 
a lower divergence indicates a lower discrepancy $\mathcal{L}^*_{\text{CFM}}$ and, consequently, a lower risk of discrepancy-induced degradation.
This quantity is computable entirely from the pre-trained model $u_\theta$ and serves as a data-free, inference-time surrogate for discrepancy.
To compute this divergence efficiently, we adopt Hutchinson's trace estimator~\cite{hutchinson89hutchinson}, following its successful application in the literature~\cite{grathwohl2018ffjord}.
The reliability of this surrogate is further discussed and evaluated in App.~\ref{app:reliability}. 

\vspace{-6pt}
\subsection{Flow Divergence Sampler}
\label{sec:method_fds}
\vspace{-3pt}
Building upon the theoretical foundations established in Sec.~\ref{sec:surrogate}, we introduce the Flow Divergence Sampler (FDS). 
At each inference timestep $t$, the core operational objective of FDS is to optimize the solver state $x_t$ to $\tilde{x}_t$ such that the refined state exhibits lower discrepancy.
A direct way to achieve this would be to optimize $\hat{\delta}_t(x)$ with respect to $x$ via gradient descent, yet this is computationally unattractive for large generative models.
Since $\hat{\delta}_t(x)$ already contains a first-order spatial derivative of the flow model, differentiating it with respect to $x$ would require second-order derivatives, which are expensive in high-dimensional settings.

To circumvent this computational overhead, FDS performs a \emph{zero-order local refinement} by applying small random perturbations, a mechanism inspired by \cite{ma2025sop}. 
These perturbations are simply scaled by a decaying schedule $\sigma_t$, which gradually decreases from $\sigma_{max}$ to $0$. 
At a solver state $x_t$, we construct a set of nearby $M$ candidates:
\begin{equation}
    x^{(0)}=x_t, ~~~x^{(m)}=x_t + \sigma_t\xi^{(m)},~~~\xi^{(m)}\sim\mathcal{N}(0,I), ~~~m=1,...,M.
\end{equation}
Then for each candidate, we evaluate its discrepancy using \cref{eqn:proxy}, and select the best candidate,
\begin{equation}
    m^* = \underset{m \in \{0, \dots, M\}}{\arg\min} \hat{\delta}_t (x^{(m)}),~~~~~\tilde{x}_t\leftarrow x^{(m^*)},
\end{equation}
and repeat this procedure for $N$ refinement iterations, using the updated $\tilde{x}_t$ as the base state $x^{(0)}$ if desired. 
The resulting state is denoted by $\tilde{x}_t$.

As demonstrated, unlike conventional advanced samplers, the key distinction of FDS is that it performs spatial corrections on the state toward a reliable region.
Standard high-order ODE solvers primarily improve generation by reducing temporal integration error once the current state $x_t$ is fixed. 
In contrast, FDS performs a spatial intervention at a fixed time $t$, relocating the intermediate state itself before the next solver step is taken. 
Because this refinement is applied externally to the base solver, FDS can be attached to Euler, Heun, and other off-the-shelf solvers. 

One sampling step with FDS therefore takes the form,
\begin{equation}
\label{eq:refinesolverstep}
    \tilde{x}_{t_k}=\text{Refine}({x}_{t_k}, t_k; u_\theta),~~~{x}_{t_{k+1}}=\text{SolverStep}(\tilde{x}_{t_k},t_k, t_{k+1};u_\theta).
\end{equation}
Thus, FDS is a plug-in inference-time module—readily compatible with various base solvers and guidance methods—rather than their replacement.
A psuedo-code of our framework is provided in the App.~\ref{app:fds_algo}.

\vspace{-5pt}
\subsubsection{Computational Overhead}
While FDS introduces extra computations to the base solver proportional to the number of refinement iterations $N$ and the number of search candidate $M$, empirical evaluations reveal a crucial practical insight: successfully bypassing high-discrepancy regions does not demand an expensive computational budget. 
In practice, we found that setting $N{=}M{=}1$ is sufficient to significantly elevate generation fidelity compared to baseline models with more NFEs, as demonstrated by the extensive evaluations in Sec.~\ref{sec:class_cond}.

To further maximize efficiency, we explicitly deactivate FDS in the early stages of generation ($t < T_{\text{trunc}}$), empirically setting the truncation threshold $T_{\text{trunc}} = 0.5$. 
Since severe trajectory conflicts predominantly manifest during these earlier denoising stages~\cite{bertrand2025closedformfm}, this temporal constraint ensures optimal resource allocation without much sacrificing quality. 
Collectively, these spatial and temporal design choices establish FDS as a low-overhead, highly practical inference-time solution. 
An in-depth analysis of these components—including the choices for the temporal boundary ($T_{\text{trunc}}$) and $N$, $M$—is provided in our ablation study (Sec. \ref{sec:ablation}).

\vspace{-6pt}
\section{Related Works}
\vspace{-2pt}
\subsection{Resolving Crossings on Train Time}
\label{sec:related_work_training}
\vspace{-2pt}
Recent studies increasingly highlight the limitations induced by intersecting trajectories in flow matching. To resolve this, some methods fundamentally change the formulation to allow for crossings; HRF \cite{zhang2025hrm} and VRFM \cite{guo2025vrm} reformulate the framework to capture the full multi-modal velocity distribution (Fig.~\ref{fig:crossing}), and CAF \cite{park2024caf} utilizes an acceleration-based approach to navigate away from empty data regions. Conversely, other works \cite{lee2023trajcurve, kim2024sfno} attempt to straighten the trajectory by introducing auxiliary encoders. 
While these strategies successfully mitigate trajectory crossings, their strict reliance on fundamental architectural changes and prohibitively expensive from-scratch training renders them highly inefficient, strongly motivating the need for a training-free alternative.

\vspace{-5pt}
\subsection{Enhancing Generation Quality at Inference-Time}
\vspace{-2pt}
To boost generative fidelity and image quality without requiring additional training, recent literature has explored advanced inference solvers to reduce temporal error. 
Foundational methods~\cite{liu2022pndm, lu2022dpm} employ higher-order numerical approximations to follow trajectory curvature, and techniques like UniPC~\cite{zhao2023unipc} utilize history buffers to optimize this process for few-step generation. 
Additionally, recent advanced samplers~\cite{luo2026lookahead} have been proposed to dynamically allocate computational resources by evaluating trajectory stiffness. 
However, all these approaches primarily focus on mitigating numerical discretization errors, restricting them to temporal adjustments along a fixed integration path. Our work fundamentally differs by providing spatial enhancements, actively steering the generation away from severe trajectory crossings.

Another line of research focuses on advanced guidance mechanisms that spatially update the state $x_t$ at inference time. 
For instance, recent studies~\cite{bai2024zigzag, wang2025goldencfg} formulate objectives composed with conditional and unconditional velocities, but are primarily restricted to classifier-free guidance~\cite{ho2022cfg} settings. Similarly, other works~\cite{he2024manifold, ye2024tfg} refine $x_t$ using external reward models or predefined objectives, inherently requiring external signals. In contrast, our approach operates as a versatile, plug-and-play module that requires no external signals and seamlessly integrates with existing inference-time enhancements.

\section{Experiments} \label{sec:exp}
We validate the effectiveness of FDS across a variety of settings.
We first illustrate the intuition and underlying mechanism of FDS on 2D synthetic data in Sec.~\ref{sec:toy_exp}.
Furthermore, we mainly validate FDS on ImageNet $256\times256$~\cite{deng2009imagenet} and CIFAR-10~\cite{krizhevsky2009cifar10} in Sec.~\ref{sec:class_cond}, demonstrating its consistent robustness across diverse backbones and solvers.
Finally, we highlight the plug-and-play versatility of FDS in diverse tasks such as text-to-image synthesis and inverse problems in Sec.~\ref{sec:fds_app}.
Detailed setups for each experiment are placed in the App.~\ref{app:implementation}.

\begin{figure}[!t]
  \centering

  \begin{subfigure}[t]{0.99\linewidth}
    \centering
    \includegraphics[width=\linewidth]{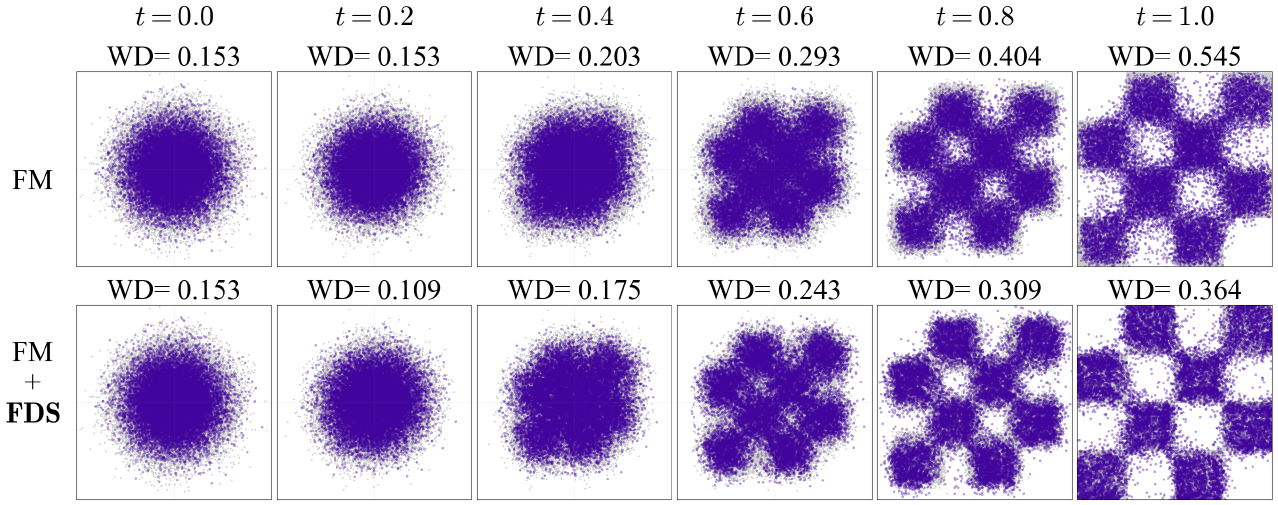}
    \caption{
  FDS yields improved generation quality, reflected by a lower Wasserstein Distance (WD).
    }
    \label{fig:toy_a}
  \end{subfigure}

  \vspace{1pt}

  \begin{subfigure}[t]{0.48\linewidth}
    \centering
    \includegraphics[width=\linewidth]{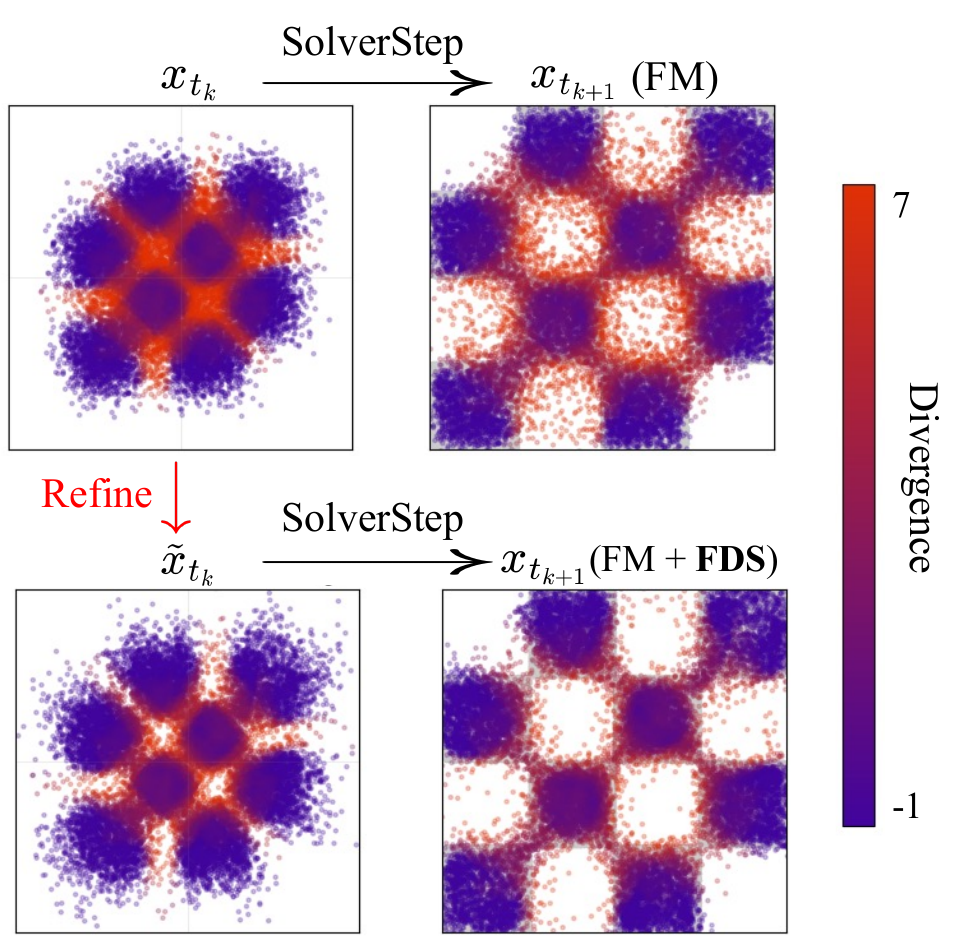}
    \caption{FDS \textit{refines} $x_t$ into $\tilde{x}_t$ by steering it toward lower-divergence regions.}
    \label{fig:toy_b}
  \end{subfigure}\hfill
  \begin{subfigure}[t]{0.5\linewidth}
    \centering
    \includegraphics[width=\linewidth]{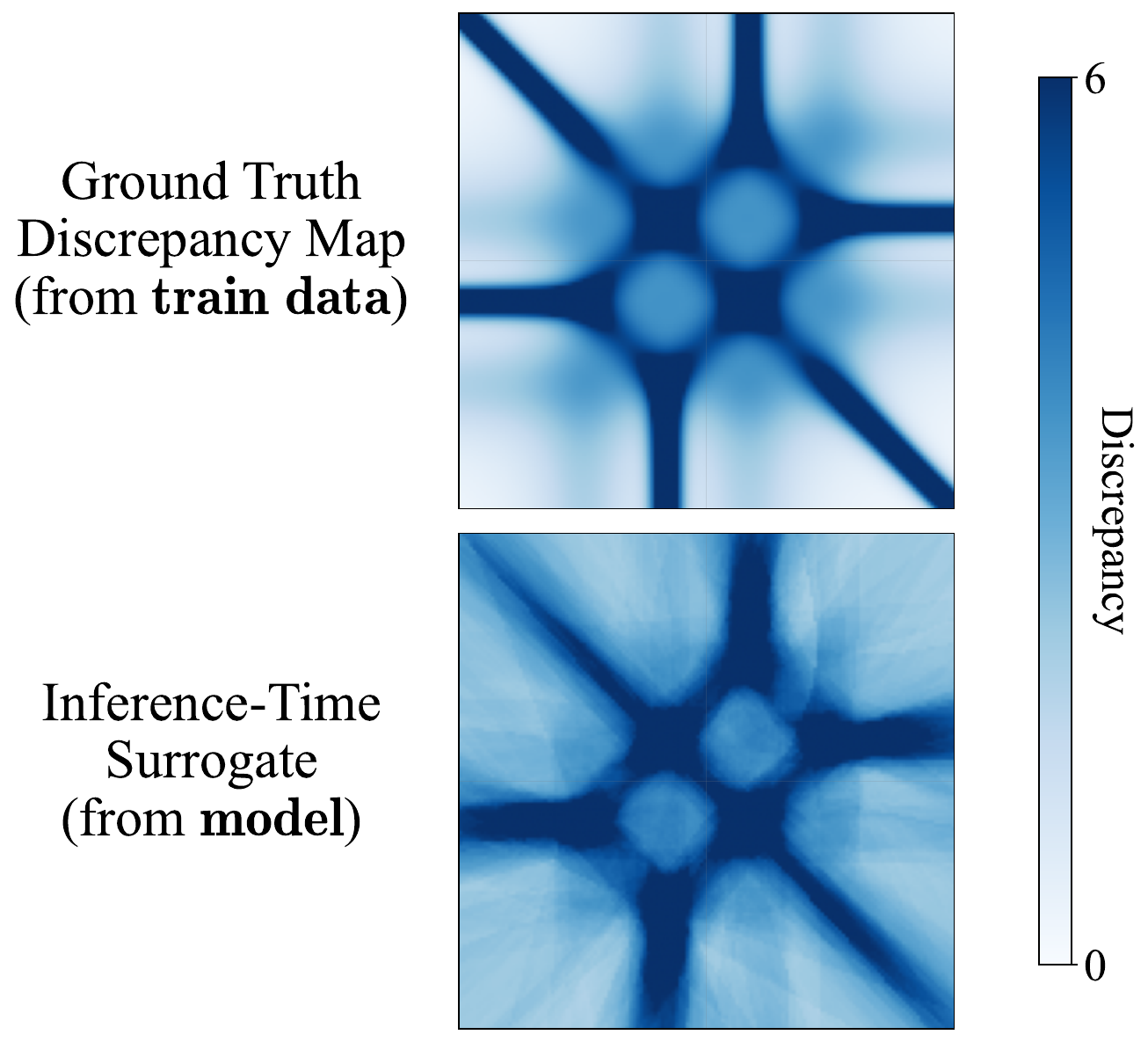}
    \caption{The discrepancy map measured by full train data (GT) and  the model at $t=0.6$.}
    \label{fig:toy_c}
  \end{subfigure}
  \vspace{-0.4em}

  \caption{\textbf{2D Synthetic Experiment} shows our divergence-based criterion correlates with sample quality. \textbf{(a)} FDS achieves more accurate modeling of the target distribution than standard FM, yielding a lower Wasserstein Distance (WD). \textbf{(b)} Standard FM passes $x_{t_k}$ directly to the ODE solver, whereas FDS refines $x_{t_k}$ into $\tilde{x}_{t_k}$, moving it to a low-divergence region. \textbf{(c)} Discrepancy maps computed from the ground-truth sample-wise velocities (Top) and from our inference-time surrogate using the pre-trained model (Bottom) are highly consistent.}
  \label{fig:toy}
  \vspace{-1.1em}
\end{figure}

\subsection{Synthetic 2D Experiments}
\label{sec:toy_exp}
\vspace{-3pt}

To intuitively illustrate the core mechanism of FDS, we first conduct a synthetic experiment mapping a standard Gaussian prior to a 2D checkerboard distribution.
Fig.~\ref{fig:toy_a} visualizes the generation processes of standard Flow Matching (FM) with an Euler solver alongside our FDS.
We quantitatively assess the generation quality using the Wasserstein Distance (WD) to the exact path, reported above each snapshot.
As $t$ progresses, FDS consistently achieves a lower WD compared to standard FM.
Notably, at $t=1.0$, FDS significantly reduces the number of stray samples outside the target regions, successfully mitigating quality degradation and ensuring high-fidelity generation.

\subsubsection{Understanding Behavior of FDS}
To further investigate the mechanism, we visualize the intermediate refinement process at $t_k=0.6$ in Fig.~\ref{fig:toy_b}.
All samples are color-coded by their initial divergence at $x_{t_k}$.
As shown, FDS effectively steers high-divergence samples (orange) toward low-divergence regions (purple).
Following the solver step to $t_{k+1}=1.0$, the unrefined orange samples stray off the checkerboard.
In contrast, the refined samples correctly align with the target distribution.
Collectively, these observations demonstrate that our enhanced generation quality directly stems from mitigating the discrepancy between marginal and sample-wise velocities on high-divergence region.

\subsubsection{Divergence as a Reliable Surrogate}
While empirical results support that our proposed surrogate (\cref{eqn:proxy}) serves as a highly reliable discrepancy indicator during inference, we further validate its accuracy by comparing the true discrepancy (\cref{eq:optimal_objective}) with our estimation (\cref{eq:main_div}) measured with the surrogate.
To this end, we visualize both discrepancy maps in Fig.~\ref{fig:toy_c}.
The top panel illustrates the true discrepancy explicitly measured by the training data distribution, whereas the bottom panel depicts our data-free estimation via flow model divergence.
The strong structural alignment between the two maps implies that our divergence-based estimation effectively captures the underlying true discrepancy, thereby demonstrating the empirical soundness of our approach.

\subsection{Main Experiments}
\label{sec:class_cond}

\begin{figure}[!t]
\vspace{-6pt}
\centering

\begin{minipage}{\textwidth}
\centering
\scriptsize
\setlength{\tabcolsep}{3.2pt}
\renewcommand{\arraystretch}{1.08}

\captionof{table}{\textbf{Performance comparison on CIFAR-10 and ImageNet $256\times256$}.
FDS consistently improves generation quality in terms of FID across all configurations in the main experiments.
$\dagger$ denotes a base solver with an increased NFEs to match the wall-clock time of our framework.}
\label{tab:main}
\vspace{7pt}

\begin{tabular}{lccccccccc}
\toprule
                        &   & \multicolumn{4}{c}{\textbf{CIFAR-10}} & \multicolumn{4}{c}{\textbf{ImageNet $256\times 256$}} \\
\cmidrule(lr){3-6} \cmidrule(lr){7-10}
\multicolumn{1}{c}{\textbf{Solver}} &\multicolumn{1}{c}{\textbf{NFE}}  & \multicolumn{2}{c}{\textbf{EDM Cond.}} & \multicolumn{2}{c}{\textbf{EDM Uncond.}} &  \multicolumn{2}{c}{\textbf{JiT-B/16}} & \multicolumn{2}{c}{\textbf{JiT-L/16}} \\
\cmidrule(lr){3-4} \cmidrule(lr){5-6}\cmidrule(lr){7-8} \cmidrule(lr){9-10}
                           & & FID ($\downarrow$)  & IS ($\uparrow$)  & FID ($\downarrow$)  & IS ($\uparrow$) &  FID $\downarrow$) & IS ($\uparrow$) & FID ($\downarrow$) & IS ($\uparrow$) \\
\midrule
Euler               & 50      & 3.003              & 9.576               & 3.034     & 9.371    & 4.151            & 280.07             & 3.859         & 277.09  \\
Euler$^\dagger$     & 77      & 2.515              & \textbf{9.671}       & 2.550     & \textbf{9.464} & 4.061        & \textbf{287.15}     & 3.857         & \textbf{278.75} \\
\arrayrulecolor[HTML]{C0C0C0}\midrule
\arrayrulecolor{black}
Euler + \textbf{FDS}& 50      & \textbf{2.319}     & 9.660               & \textbf{2.440} & 9.387 & \textbf{3.799} & 278.33 & \textbf{3.519} & 278.16 \\
\midrule
Heun                & 99      & 1.904              & 9.829               & 2.021     & 9.615    & 3.637            & 270.18             & 2.713         & 330.23  \\
Heun$^\dagger$      & 153      & 1.910              & 9.820               & 2.034     & 9.606    & 3.815            & \textbf{275.10}    & 2.886         & \textbf{333.10} \\
\arrayrulecolor[HTML]{C0C0C0}\midrule
\arrayrulecolor{black}
Heun + \textbf{FDS} & 99      & \textbf{1.786}     & \textbf{9.875}      & \textbf{1.953} & \textbf{9.641} & \textbf{3.394} & 269.09 & \textbf{2.496} & 329.70 \\
\bottomrule
\end{tabular}
\end{minipage}

\vspace{6.5pt} 

\begin{minipage}{\textwidth}
\centering
\includegraphics[width=0.99\linewidth]{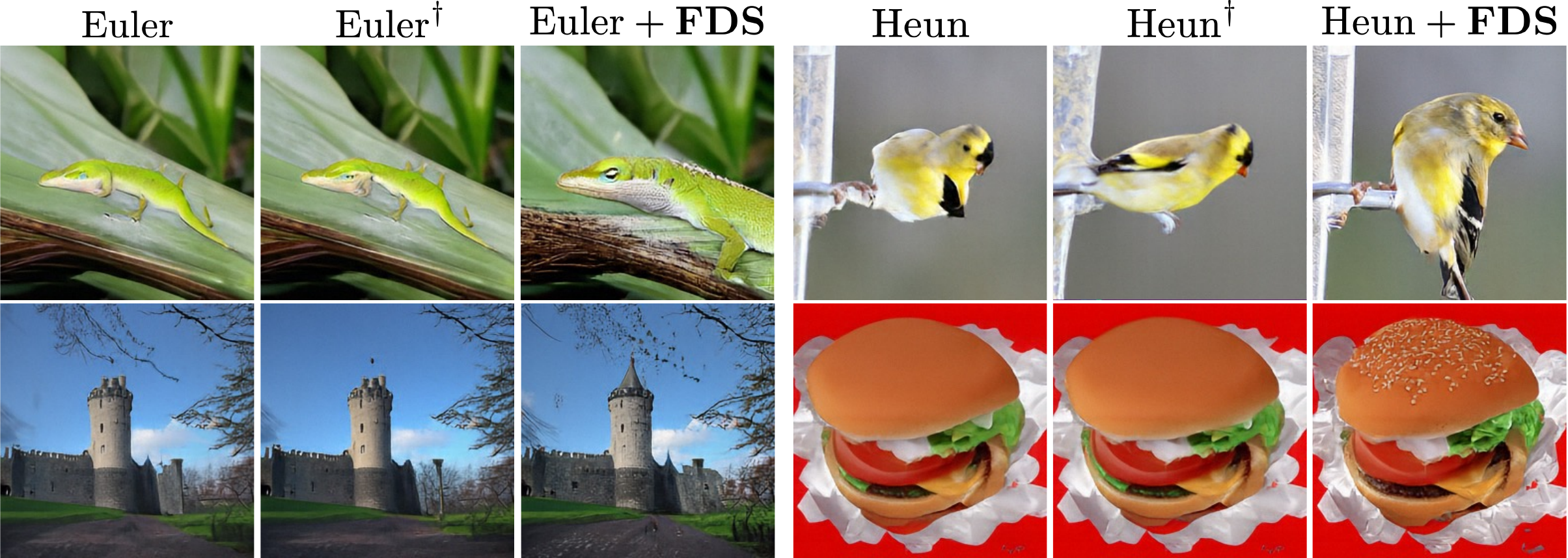}
\vspace{-4pt}
\captionof{figure}{\textbf{Qualitative results on ImageNet} $\mathbf{256\times 256}$ with JiT-L/16.
Compared to the compute-matched baseline($\dagger$), FDS effectively enhances the generation quality. 
}
\label{fig:qual_samples}
\end{minipage}

\vspace{-8pt}
\end{figure}

\subsubsection{Experimental Setup}
To validate the applicability of FDS on pre-trained models, we conduct experiments on CIFAR-10~\cite{krizhevsky2009cifar10} and ImageNet $256\times 256$~\cite{deng2009imagenet}.
We apply FDS to well-established flow-based models: EDM \cite{karras2022edm} for CIFAR-10 and JiT \cite{li2026jit} for ImageNet.
To evaluate the efficacy of FDS, we construct compute-matched baselines in Tab.~\ref{tab:main} (marked with $^\dagger$) 
by allocating additional steps to match the wall-clock time required by FDS. This ensures a fair comparison under an equivalent computational budget. The exact computing procedure is detailed in App.~\ref{app:implementation} and Tab.~\ref{tab:throughput}, reflecting a roughly 1.5$\times$ increase in wall-clock time for FDS on the same NFEs.
For evaluation, we report Fréchet Inception Distance (FID)~\cite{heusel2017fid} and Inception Score (IS)~\cite{salimans2016is}.

\subsubsection{Results}

As detailed in Tab.~\ref{tab:main}, FDS consistently improves the generation quality across various datasets, models, and ODE solvers, in terms of FID, demonstrating the robust applicability of our framework.
For compute-matched baselines, while increasing integration steps generally enhances performance, these gains quickly saturate with advanced solvers like Heun. 
In some cases, such as ImageNet $256 \times 256$ with the Heun solver, this even leads to performance stagnation or diminishing returns. Conversely, allocating this same computational budget to FDS yields substantial enhancements in image fidelity across all settings.

As discussed in Sec.~\ref{sec:method_fds}, this performance gap arises because FDS operates on a fundamentally different principle. 
Rather than solely relying on finer temporal integration, our method introduces targeted spatial updates to the generation process at fixed timestep $t$. 
By steering generative trajectories away from areas of high-discrepancy region, FDS effectively prevents the model from traversing regions that typically cause degraded or blurry outputs. 
Fig.~\ref{fig:qual_samples} (right) directly illustrates this advantage: while simply increasing the NFE yields only subtle localized changes, FDS successfully synthesizes fine-grained details—such as distinct sesame seeds on a hamburger—that a naive high-NFE baseline cannot produce. 
Ultimately, this active spatial refinement at inference-time achieves consistent improvements in fidelity that remain unattainable through the temporal scaling of existing samplers. 
Additional experimental results, including those on SiT~\cite{ma2024sit}, can be found in App.~\ref{app:more_exps}, while further qualitative results are presented in App.~\ref{app:more_qual}.

\begin{figure}[!t] 
\centering

\captionsetup{type=table}
\caption{\textbf{Quantitative results for text-to-image synthesis} on the DrawBench dataset. FDS consistently improves upon the baselines across most metrics, demonstrating its robustness across different model backbones.
}
\label{tab:t2i}
\vspace{-4.8pt}

\begin{subtable}[t]{0.48\linewidth}
\centering
\caption{Experiment results on SD3-M}
\label{tab:t2i_sd3m}

{\fontsize{7.7pt}{8.5pt}\selectfont
\setlength{\tabcolsep}{2.5pt}
\renewcommand{\arraystretch}{1.05}
\begin{tabular}{l cccc}
\toprule
\textbf{Method} & IR & HPSv2 & Aes. & CLIP \\
\midrule
SD3-M              & \num{82.36} & \num{27.72} & \num{5.70} & \num{28.47} \\
SD3-M+\textbf{FDS} & \textbf{\num{89.33}} & \textbf{\num{27.76}} & \textbf{\num{5.72}} & \textbf{\num{28.76}} \\
\bottomrule
\end{tabular}
}
\end{subtable}
\hfill
\begin{subtable}[t]{0.48\linewidth}
\centering
\caption{Experiment results on FLUX.1}
\label{tab:t2i_flux}

{\fontsize{7.7pt}{8.5pt}\selectfont
\setlength{\tabcolsep}{2.5pt}
\renewcommand{\arraystretch}{1.05}
\begin{tabular}{l cccc}
\toprule
\textbf{Method} & IR & HPSv2 & Aes. & CLIP \\
\midrule
FLUX.1              & \num{92.95} & \num{29.37} & \textbf{\num{6.18}} & \num{27.33} \\
FLUX.1+\textbf{FDS} & \textbf{\num{94.63}} & \textbf{\num{29.39}} & \num{6.14} & \textbf{\num{27.46}} \\
\bottomrule
\end{tabular}
}
\end{subtable}

\vspace{8pt}

\captionsetup{type=figure}
\includegraphics[width=0.96\linewidth]{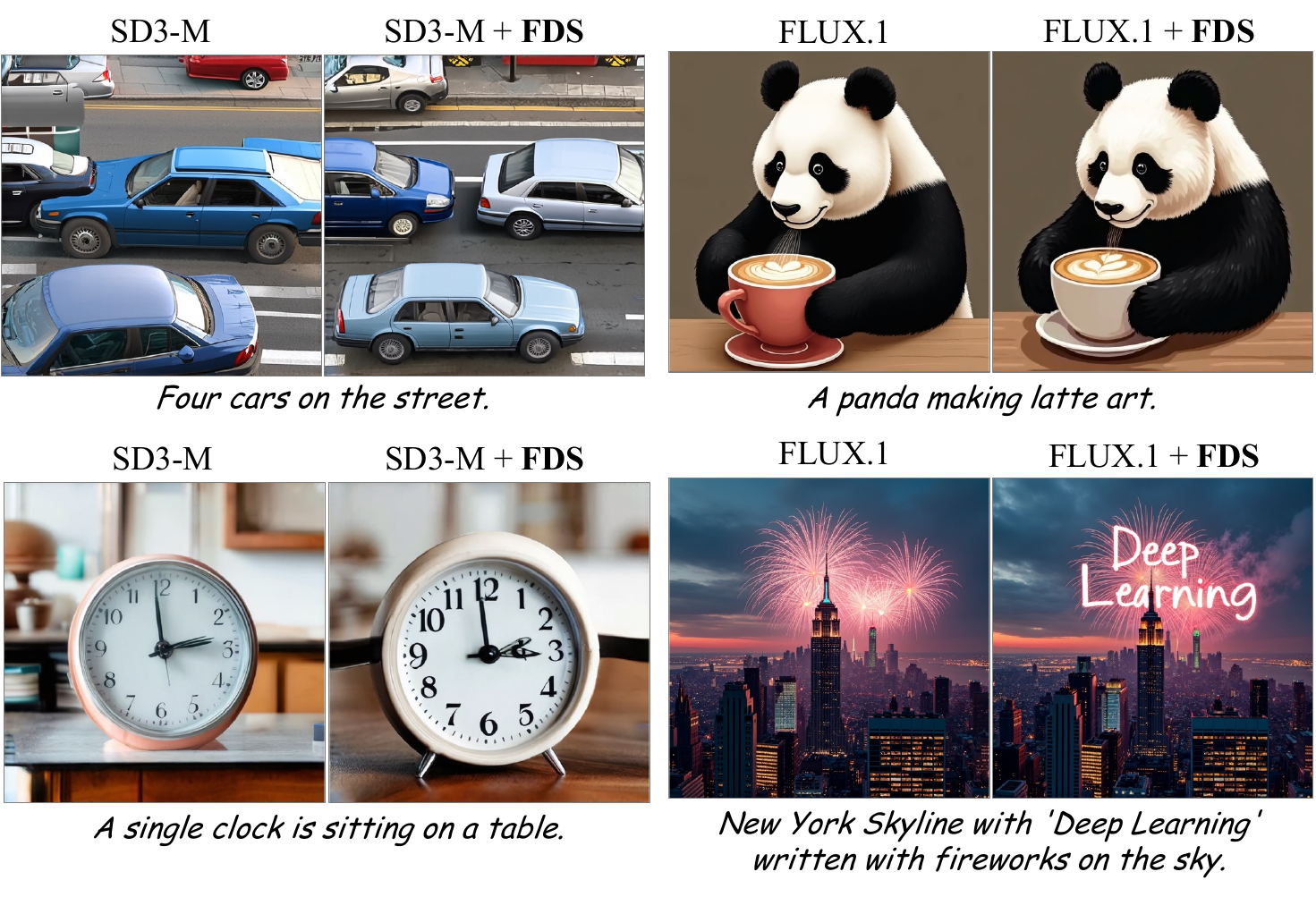}
\vspace{-1.1em}
\caption{\textbf{Qualitative results on text-to-image synthesis.} 
Images generated with FDS exhibit improved overall quality, characterized by better spatial coherence. More qualitative results can be found in App.~\ref{app:more_qual}.}
\vspace{-12pt}

\label{fig:t2i_qual_samples}

\end{figure}

\vspace{-2pt}
\subsection{FDS on Various Tasks}
\label{sec:fds_app}
\vspace{-1pt}
To further validate the robustness of our framework, we provide additional quantitative and qualitative analyses that extend beyond our primary evaluations. Specifically, we explore the versatility of FDS across diverse yet widely adopted guidance settings. This includes evaluating its behavior in text-to-image generation under varying guidance scales, testing its compatibility with standard test-time guidance mechanisms frequently utilized for inverse problems, and assessing its integration with other advanced inference-time methods. 
The following evaluations demonstrate that FDS can be seamlessly plugged into various generation pipelines, consistently maintaining high fidelity across different configurations
%

\subsubsection{Text-to-Image Generation}

We evaluate our framework on two different text-to-image generation backbones, SD3-Medium~\cite{esser2024sd3} and FLUX.1~\cite{BlackForestLabs2024FLUX1}, using the DrawBench~\cite{saharia2022drawbench} dataset. 
To comprehensively assess the overall sample quality, we utilize four established reward models widely adopted in text-to-image generation tasks. 
Specifically, we measure human preference using ImageReward (IR)~\cite{xu2023imagereward} and HPSv2~\cite{wu2023hpsv2}, evaluate visual appeal via the Aesthetic Predictor~\cite{schuhmann2022aesthetic}, and quantify text-image alignment using the CLIP Score~\cite{hessel2022clipscore}.

Quantitative results in Tab.~\ref{tab:t2i} and qualitative samples indicate that FDS can be effectively applied to text-to-image generation tasks, offering improvements in visual quality and showing robustness across different backbones.
Specifically, integrating our framework helps produce images with enhanced fidelity, as demonstrated by the accurate rendering of fine-grained details like the clock faces and numbers in Fig.~\ref{fig:t2i_qual_samples} (bottom left). 
Overall, these enhancements support that FDS is a beneficial inference-time module that consistently boosts the performance of advanced text-to-image models.
Additional quantitative results for text-to-image generation, including experiments with varying backbone architectures and evaluations on the MS-COCO~\cite{lin2014mscoco} dataset, are provided in App.~\ref{app:more_exps}.



\vspace{-2pt}

\subsubsection{Inverse Problem}
We further demonstrate the plug-and-play capability of FDS by integrating it with inference-time guidance frameworks to enhance generation quality.
We assess this validation on image deblurring and super-resolution tasks following the setup of TFG~\cite{ye2024tfg} using the Cat~\cite{Elson2007cat} dataset, with detailed settings provided in the App.~\ref{app:implementation}.

\begin{figure*}[!t]
\vspace{-6pt}
\centering

\begin{minipage}[t]{0.41\textwidth}
  \centering
  \footnotesize
  \vspace{-55pt}
  \setlength{\tabcolsep}{2.2pt}
  \renewcommand{\arraystretch}{1.12}

  \vspace{-73pt}
  
  \captionof{table}{\textbf{Quantitative evaluation on inverse problems.} Baselines equipped with FDS demonstrate consistent improvements across all metrics. All reported values are lower-is-better.}
  \label{tab:inverse_prob}
  \vspace{4pt}

  \begin{tabular}{llcc}
    \toprule
    \textbf{Task} & \textbf{Method} & \textbf{FID} & \textbf{LPIPS} \\
    \midrule
    \multirow{2}{*}{Deblur}
      & TFG        & \num{64.02} & \num{15.50} \\
      & TFG + \textbf{FDS} & \textbf{\num{63.17}} & \textbf{\num{14.93}} \\
    \midrule
    \multirow{2}{*}{SR $\times$4}
      & TFG        & \num{65.54} & \num{18.70} \\
      & TFG + \textbf{FDS}   & \textbf{\num{63.14}} & \textbf{\num{16.23}} \\
    \bottomrule
  \end{tabular}
\end{minipage}\hfill
\begin{minipage}[t]{0.54\textwidth}
  \centering
  \includegraphics[width=0.9\linewidth]{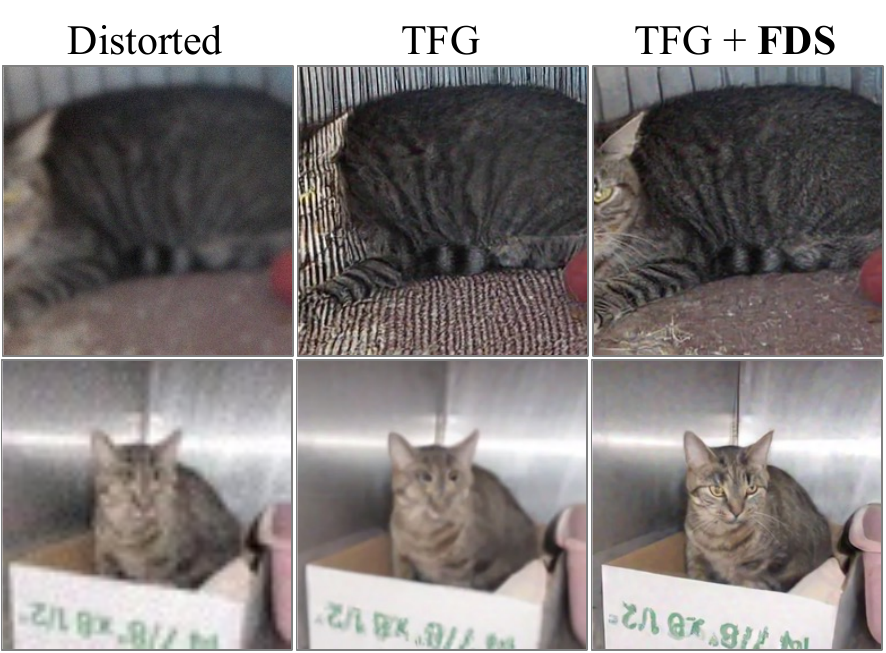}
  \vspace{-2pt}
  \captionof{figure}{\textbf{Qualitative results on inverse problems.} (Top) Deblurring. (Bottom) Super-resolution ($\times$4).}
  \label{fig:inverse_qual}
\end{minipage}

\vspace{-6pt}
\end{figure*}

\begin{figure}[t]
\centering
\footnotesize
\captionsetup{type=table}
\caption{
\textbf{FDS on diverse inference-time methods.} We compare the performance of FDS when integrated with (a) an advanced solver, (b) a higher-order solver, and (c) an advanced CFG framework. The results consistently demonstrate the effectiveness of FDS across all configurations.
}
\begin{subtable}[t]{0.31\linewidth}
\centering
\setlength{\tabcolsep}{1.2pt}
\scriptsize
\caption{FDS on advanced solver.}
\label{tab:unipc_fds}
\vspace{1pt}
\begin{tabular}{lcc}
\toprule
\textbf{Method} & \textbf{FID}($\downarrow$) & \textbf{IS}($\uparrow$) \\
\midrule
UniPC \cite{zhao2023unipc} & 6.21 & 270.8 \\
UniPC+\textbf{FDS} & \textbf{5.59} & \textbf{272.4} \\
\bottomrule
\end{tabular}
\end{subtable}
\hfill
\begin{subtable}[t]{0.31\linewidth}
\centering
\setlength{\tabcolsep}{1.2pt}
\scriptsize
\caption{FDS on high-order solver.}
\label{tab:rk4_fds}
\vspace{1pt}
\begin{tabular}{lcc}
\toprule
\textbf{Method} & \textbf{FID}($\downarrow$) & \textbf{IS}($\uparrow$) \\
\midrule
RK4 & 3.77 & \textbf{264.40} \\
RK4+\textbf{FDS} & \textbf{3.45} & 261.9 \\
\bottomrule
\end{tabular}
\end{subtable}
\hfill
\begin{subtable}[t]{0.33\linewidth}
\centering
\setlength{\tabcolsep}{1.1pt}
\scriptsize
\caption{FDS on advanced CFG.}
\label{tab:cfgzero_fds}
\vspace{1pt}
\begin{tabular}{lcc}
\toprule
\textbf{Method} & \textbf{FID}($\downarrow$) & \textbf{IS}($\uparrow$) \\
\midrule
CFG-Zero* \cite{fan2025cfgzero} & 4.10 & \textbf{276.8} \\
CFG-Zero*+\textbf{FDS} & \textbf{3.78} & 276.5 \\
\bottomrule
\end{tabular}
\end{subtable}

\vspace{-2pt}
\label{fig:fds_ablation_tables}

\vspace{-4pt}
\end{figure}

As demonstrated quantitatively in Tab.~\ref{tab:inverse_prob}, FDS consistently improves upon the baseline by generating substantially more detailed and realistic images. 
Notably, it achieves lower FID and LPIPS~\cite{zhang2018lpips}, indicating a superior capacity to capture both global structures and fine-grained details. This quantitative superiority is directly corroborated by the visual comparisons in Fig.~\ref{fig:inverse_qual}. 
While the baseline often yields blurry or less-realistic samples, our proposed framework successfully retains intricate detailed features. 
This is clearly evidenced by the rendering of the cat, where fine details such as eye pupils, complex fur textures, and individual whiskers are preserved with enhanced fidelity.

\vspace{-2pt}

\subsubsection{FDS on Advanced Inference-time Methods}

To comprehensively validate the broad compatibility and plug-and-play capability of FDS, we integrate it into several established methodologies following the settings in Sec.~\ref{sec:class_cond}. 
We empirically evaluate this adaptability across distinct generative paradigms: an advanced solver (UniPC~\cite{zhao2023unipc}, 20 NFEs), a high-order solver (RK4), and an advanced classifier-free guidance technique (CFG-Zero*~\cite{fan2025cfgzero}). As demonstrated in Tabs.~\ref{tab:unipc_fds}, \ref{tab:rk4_fds}, and \ref{tab:cfgzero_fds}, our framework consistently enhances generation quality in every scenario. 
These results demonstrate that rather than competing with existing enhancements, FDS serves as a highly compatible module capable of seamlessly elevating performance across diverse inference-time settings.


\vspace{-5pt}
\section{Analysis}

\vspace{-2pt}
\subsection{Comparison with Training-Based Frameworks}
Besides the benefits of FDS being a training-free and plug-and-play method, we also observe that such approach can outperform training-based methods in terms of generation quality.
The comparison in Tab.~\ref{tab:traning-based} shows that FDS, when used with the EDM, surpasses the training-based method for resolving crossing (\emph{i.e.}, HRF~\cite{zhang2025hrm}) by a large margin in a parameter-count matched setup.

\begin{wraptable}{r}{0.46\linewidth}
\footnotesize
\centering
\vspace{-28pt} 
\caption{Comparison with training-based frameworks. 
For VRFM$^*$, we report the original values from the paper as the source code is not publicly available.
All configurations utilize Euler solver (50 NFEs) on CIFAR-10.
}
\vspace{1.7pt}
\setlength{\tabcolsep}{2.5pt}
\begin{tabular}{lccc}
\toprule
Model & {FID} & {IS} & Param.\\
\midrule
VRFM$^*$ \cite{guo2025vrm} & \num{5.27}& -& 37.2M \\
HRF \cite{zhang2025hrm} & \num{4.96} & \num{8.98} & 56.0M \\
EDM \cite{karras2022edm}  & \num{3.04}                  & \num{9.37} & 55.7M \\
EDM + \textbf{FDS}         & \textbf{\num{2.44}}        & \textbf{\num{9.39} }& 55.7M \\
\bottomrule
\end{tabular}
\label{tab:traning-based}
\vspace{-10pt} 
\end{wraptable}
We conjecture that the performance gap stems from the discrepancy during training induces a severe capacity bottleneck for these methods as they model the complex distribution of individual sample-wise velocities, rather than simply regressing their marginal velocity.
Therefore, the baselines demonstrate degraded performance compared to EDM, which retains the original training objective.
Since FDS bypasses this issue and operates entirely at inference-time, it can be applied to strong pre-trained flow models without any modifications or re-training.
By steering generation away from high-discrepancy regions on-the-fly, FDS can effectively resolves the discrepancy and mitigates the performance limitations in the training-based approaches, solidifying its practical advantage.


\subsection{Ablation Study} \label{sec:ablation}

To provide a deeper understanding of our proposed framework and to justify our default hyperparameter settings, we conduct comprehensive ablation studies that validate the core design choices abstracted in Sec.~\ref{sec:method_fds}.
All ablation experiments are conducted on ImageNet $256 \times 256$ with JiT-B/16 backbone and an Euler solver with 50 NFEs.





\vspace{-4pt}
\subsubsection{Efficacy of Early-Stage Interventions}

\begin{figure*}[t]
  \centering
  
  \begin{subfigure}[b]{0.31\linewidth}
    \centering
    \includegraphics[width=0.95\linewidth]{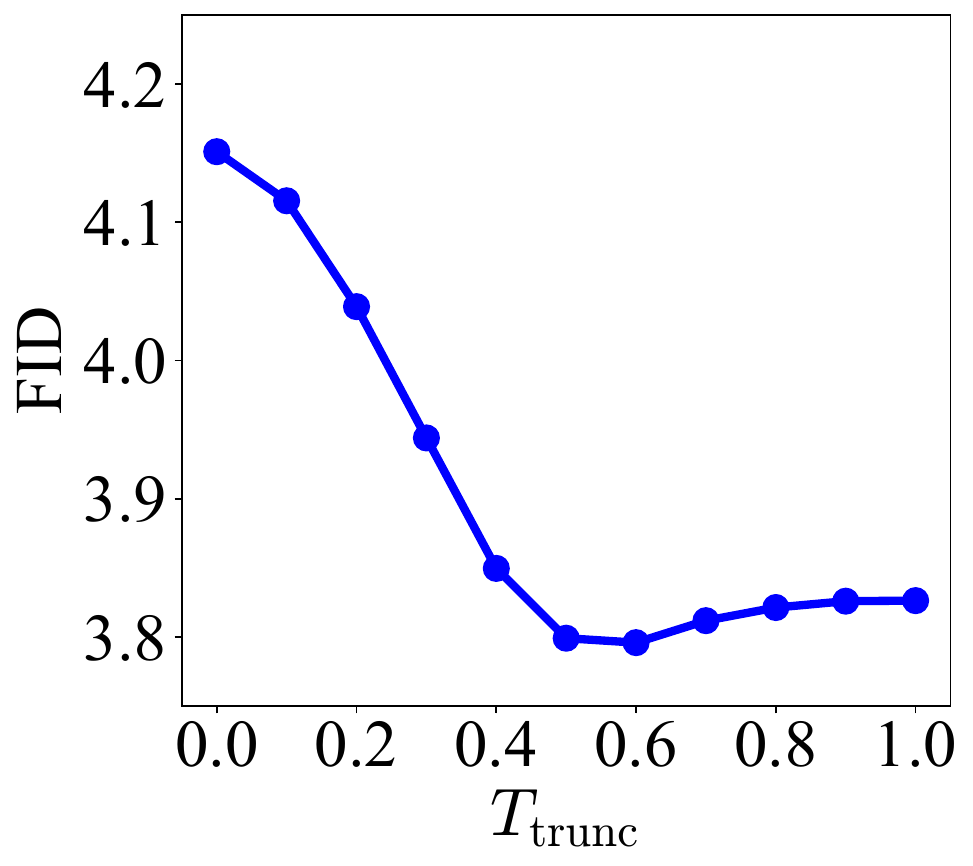}
    \caption{Ablation on $T_{\text{trunc}}$}
    \label{fig:stop_t}
  \end{subfigure}\hfill
  \begin{subfigure}[b]{0.34\linewidth}
    \centering
    \includegraphics[width=0.95\linewidth]{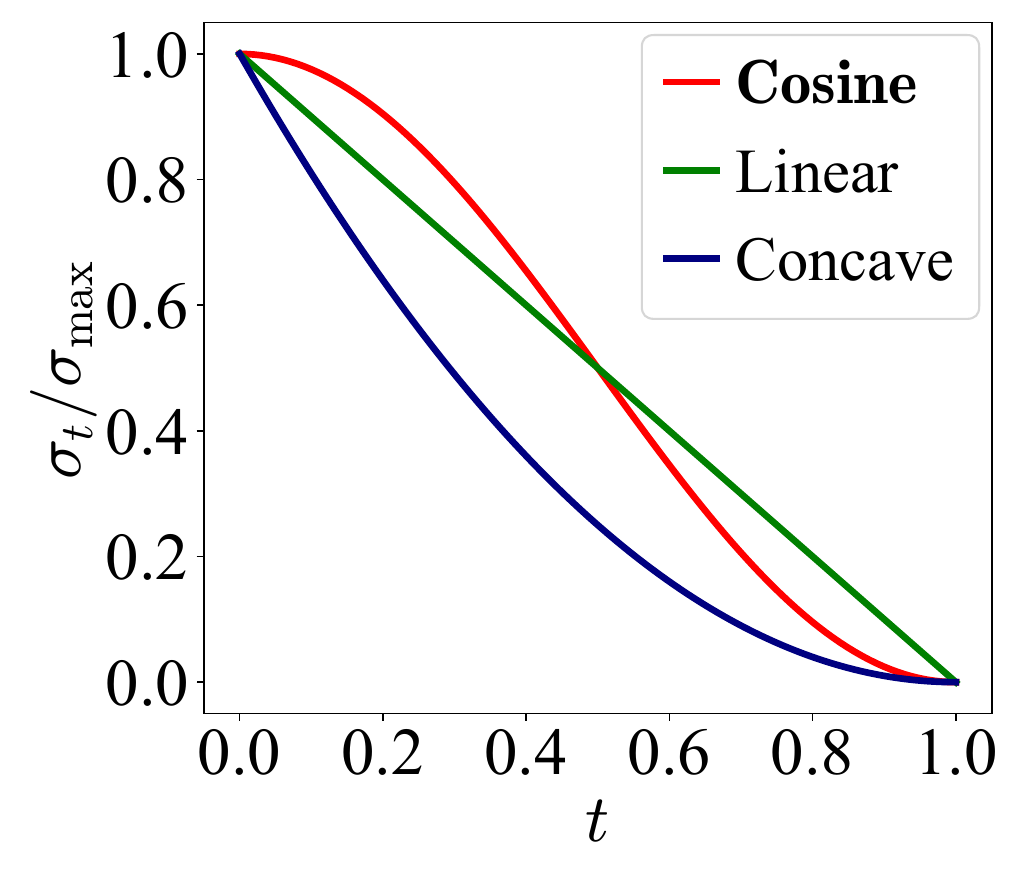}
    \vspace{-5pt}
    \caption{Visualization of $\sigma_t$ schedule}
    \label{fig:sigma_t_vis}
  \end{subfigure}\hfill
  \begin{subfigure}[b]{0.32\linewidth}
    \centering
    \footnotesize
    \setlength{\tabcolsep}{2.0pt}
    \renewcommand{\arraystretch}{1.10}
    \begin{tabular}{lc}
      \toprule
      $\sigma_t$ schedule & {FID $(\downarrow)$} \\
      \midrule
      {\color{gray} Baseline} & {\color{gray}\num{4.151}} \\
      \textbf{Cosine} & \num{3.799} \\
      Linear  & \num{3.823} \\
      Concave & \num{3.893} \\
      \bottomrule
    \end{tabular}
    \vspace{10pt} 
    \caption{Ablation on $\sigma_t$ schedule}
    \label{tab:t_schedule}
  \end{subfigure}
  \vspace{-1pt}
  \caption{\textbf{Ablation on early-stage refinement.} Applying FDS during earlier denoising stages effectively improves generation quality. \textbf{Bold text} indicates the default configuration ($T_\text{trunc}{=}0.5$, cosine schedule).
  }
  \vspace{-7pt}
  \label{fig:ablation_time}
\end{figure*}

To analyze the impact of FDS across different generation stages, we investigate restricting the refinements exclusively to timesteps $t \le T_{\text{trunc}}$.
As shown in Fig.~\ref{fig:stop_t}, the FID improves as $T_{\text{trunc}}$ increases but saturates around $T_{\text{trunc}}=0.5$.
This suggests that FDS yields the most substantial benefits during the initial stages of flow integration rather than in later phases.
Furthermore, Fig.~\ref{tab:t_schedule} indicates that adopting a cosine schedule for the perturbation scale $\sigma_t$, which allocates larger weights to earlier timesteps, surpasses both linear and concave schedules, further emphasizing the critical role of early-stage refinement.
Taken together, these findings imply that the refinements during the early phases of generation are particularly effective in steering the samples away from high-discrepancy regions.

\vspace{-3pt}

\subsubsection{Refinement Iteration $N$}

While increasing the number of iterative updates for $x_t$ yields progressive improvements in visual fidelity, we observe that the performance gains eventually plateau. 
Guided by this empirical observation (Fig. \ref{fig:iter}), we opt to prioritize computational efficiency at inference time. 
Specifically, our results indicate that a single update step is sufficient to achieve highly competitive generation quality. 
Consequently, we adopt $N=1$ and $\sigma_{\text{max}} = 0.01$ as our default experimental setting, finding that this configuration strikes a practical balance between the representational benefits of FDS and minimal computational overhead.

\subsubsection{Number of Candidates $M$}

\begin{figure*}[t]
  \centering
  \begin{subfigure}[b]{0.44\linewidth}
    \centering
    \includegraphics[width=0.95\linewidth]{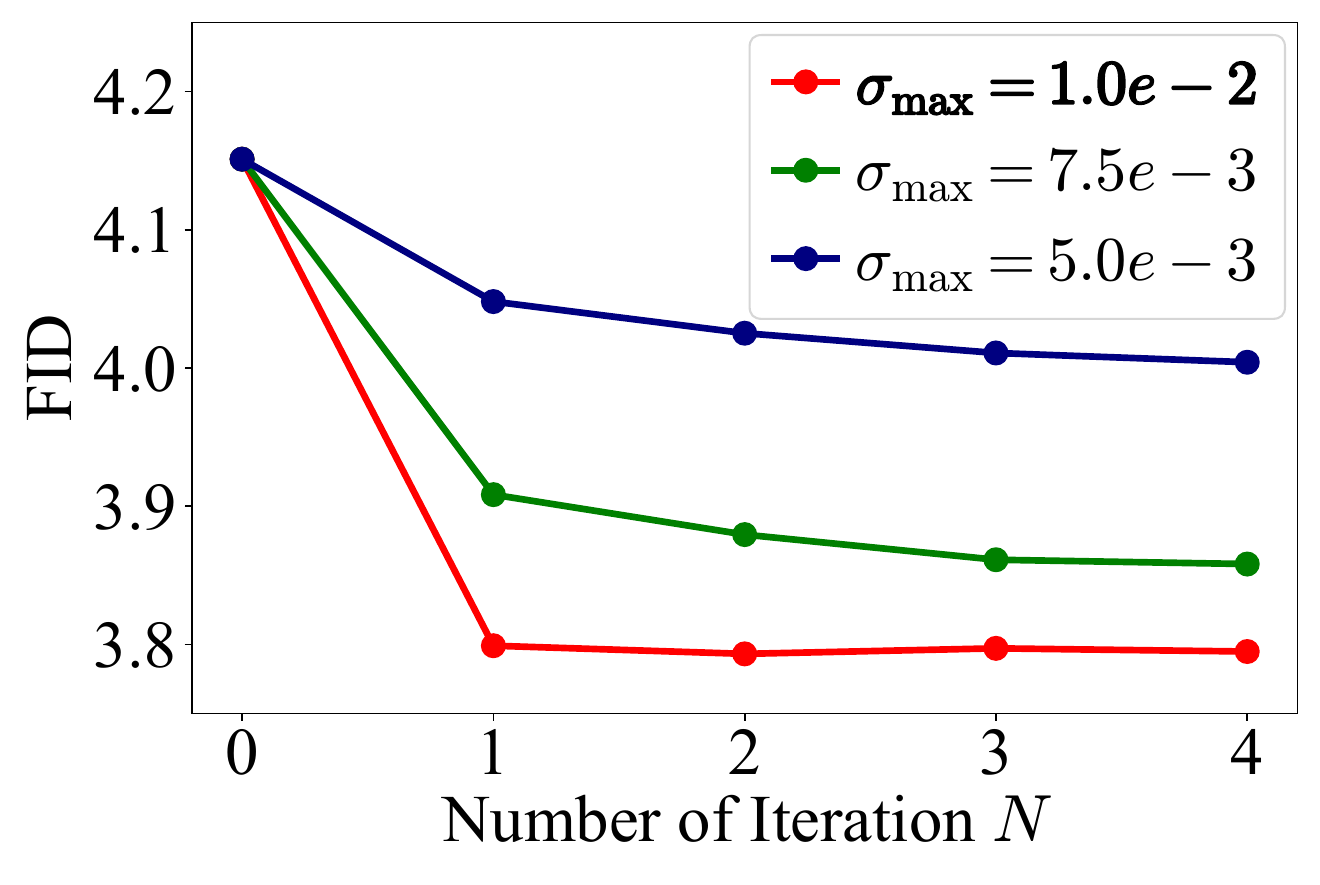}
    \caption{Ablation on number of iteration}
    \label{fig:iter}
  \end{subfigure}
  \hspace{0.05\linewidth} 
  \begin{subfigure}[b]{0.32\linewidth}
    \centering
    \footnotesize
    \setlength{\tabcolsep}{2.8pt}
    \begin{tabular}{lc}
      \toprule
      $M$ & {FID $(\downarrow)$} \\
      \midrule
      {\color{gray} 0 (Baseline)} & {\color{gray}\num{4.151}} \\
      \textbf{1}        & \num{3.799}               \\
      2  & \num{3.795} \\
      8  & \num{3.785} \\
      20 & \num{3.765} \\
      \bottomrule
    \end{tabular}
    \vspace{10pt} 
    \caption{Ablation on number of perturbation candidates}
    \label{tab:num_particles}
  \end{subfigure}

  \caption{\textbf{Effect of iterations $N$ and candidates $M$.}
    While increasing $N$ and $M$ generally improves performance, setting $N=M=1$ offers the best trade-off.}
    \vspace{-6pt}
  \label{fig:ablation_iteration}
\end{figure*}

To further explore the operational dynamics of FDS, we investigate the impact of varying the number of candidates $M$ introduced in Sec.~\ref{sec:method_fds}. 
Our empirical evaluations suggest that a single particle is generally sufficient to minimize discrepancy, offering a favorable balance between computational efficiency and generative quality. 
As shown in Fig.~\ref{tab:num_particles}, while increasing the candidate count $M$ yields subtle but consistent improvements, the overall performance quickly plateaus. 
We hypothesize that this occurs because a single perturbation, when applied consistently at every timestep, cumulatively guides the representation toward a stable, adjacent conditional mode. 
Admittedly, evaluating multiple candidates presents a distinct computational advantage: unlike increasing sequential NFEs, these evaluations can be fully parallelized without incurring additional latency overhead. 
Nevertheless, as scaling to multiple candidates yields only marginal generative benefits, the single-particle configuration remains a highly practical and efficient default choice.

\vspace{-3pt}
\section{Conclusion}
\vspace{-2pt}
In this work, we introduced a novel, training-free inference framework designed to mitigate the discrepancy between marginal and instantaneous sample-wise velocities in flow matching. 
Unlike existing solvers that primarily focus on minimizing temporal discretization errors, our approach actively steers the trajectory away from high-discrepancy regions to enhance generation quality. 
Extensive evaluations demonstrate that FDS is highly robust, yielding consistent performance improvements across various generation tasks. 
Ultimately, we believe our exploration of spatial trajectory refinement establishes a strong foundation for future research, offering a new perspective on reliable generation.


\vspace{-2pt}
\section*{Acknowledgements}
This work was in part supported by the National Research Foundation of Korea (RS-2024-00351212 and RS-2024-00436165), the Institute of Information \& communications Technology Planning \& Evaluation (IITP) (RS-2024-00509279, RS-2022-II220926, and RS-2022-II220959, RS-2019-II190075), and  the “HPC support” project funded by the Korea government. This work was supported by the InnoCORE program of the Ministry of Science and ICT (AI Meta-Scientist, N10260110).

\clearpage

%
%
\bibliographystyle{splncs04}
\bibliography{main}

@String(CVPR  = {IEEE Conf. Comput. Vis. Pattern Recog.})

@String(CVPR  = {CVPR})

@misc{esser2024sd3,
      title={Scaling Rectified Flow Transformers for High-Resolution Image Synthesis}, 
      author={Patrick Esser and Sumith Kulal and Andreas Blattmann and Rahim Entezari and Jonas Müller and Harry Saini and Yam Levi and Dominik Lorenz and Axel Sauer and Frederic Boesel and Dustin Podell and Tim Dockhorn and Zion English and Kyle Lacey and Alex Goodwin and Yannik Marek and Robin Rombach},
      year={2024}
}

@inproceedings{peebles2023dit,
  title={Scalable diffusion models with transformers},
  author={Peebles, William and Xie, Saining},
  booktitle={Proceedings of the IEEE/CVF international conference on computer vision},
  pages={4195--4205},
  year={2023}
}

@article{zhang2025hrm,
  title={Towards hierarchical rectified flow},
  author={Zhang, Yichi and Yan, Yici and Schwing, Alex and Zhao, Zhizhen},
  journal={arXiv preprint arXiv:2502.17436},
  year={2025}
}

@inproceedings{
    guo2025vrm,
    title={Variational Rectified Flow Matching},
    author={Pengsheng Guo and Alex Schwing},
    booktitle={Forty-second International Conference on Machine Learning},
    year={2025},
}

@article{grathwohl2018ffjord,
  title={Ffjord: Free-form continuous dynamics for scalable reversible generative models},
  author={Grathwohl, Will and Chen, Ricky TQ and Bettencourt, Jesse and Sutskever, Ilya and Duvenaud, David},
  journal={arXiv preprint arXiv:1810.01367},
  year={2018}
}

@misc{park2024caf,
      title={Constant Acceleration Flow}, 
      author={Dogyun Park and Sojin Lee and Sihyeon Kim and Taehoon Lee and Youngjoon Hong and Hyunwoo J. Kim},
      year={2024},
}

@misc{karras2022edm,
      year={2022},
      title={Elucidating the Design Space of Diffusion-Based Generative Models}, 
      author={Tero Karras and Miika Aittala and Timo Aila and Samuli Laine},
}

@misc{li2026jit,
      title={Back to Basics: Let Denoising Generative Models Denoise}, 
      author={Tianhong Li and Kaiming He},
      year={2026}, 
}

@misc{lipman2023fm,
      title={Flow Matching for Generative Modeling}, 
      author={Yaron Lipman and Ricky T. Q. Chen and Heli Ben-Hamu and Maximilian Nickel and Matt Le},
      year={2023}
}

@misc{zhao2023unipc,
      title={UniPC: A Unified Predictor-Corrector Framework for Fast Sampling of Diffusion Models}, 
      author={Wenliang Zhao and Lujia Bai and Yongming Rao and Jie Zhou and Jiwen Lu},
      year={2023},
}

@misc{liu2022pndm,
      title={Pseudo Numerical Methods for Diffusion Models on Manifolds}, 
      author={Luping Liu and Yi Ren and Zhijie Lin and Zhou Zhao},
      year={2022},
}

@misc{saharia2022drawbench,
      title={Photorealistic Text-to-Image Diffusion Models with Deep Language Understanding}, 
      author={Chitwan Saharia and William Chan and Saurabh Saxena and Lala Li and Jay Whang and Emily Denton and Seyed Kamyar Seyed Ghasemipour and Burcu Karagol Ayan and S. Sara Mahdavi and Rapha Gontijo Lopes and Tim Salimans and Jonathan Ho and David J Fleet and Mohammad Norouzi},
      year={2022},
}

@misc{bertrand2025closedformfm,
      title={On the Closed-Form of Flow Matching: Generalization Does Not Arise from Target Stochasticity}, 
      author={Quentin Bertrand and Anne Gagneux and Mathurin Massias and Rémi Emonet},
      year={2025},
}

@inproceedings{
    ye2024tfg,
    title={{TFG}: Unified Training-Free Guidance for Diffusion Models},
    author={Haotian Ye and Haowei Lin and Jiaqi Han and Minkai Xu and Sheng Liu and Yitao Liang and Jianzhu Ma and James Zou and Stefano Ermon},
    booktitle={The Thirty-eighth Annual Conference on Neural Information Processing Systems},
    year={2024}
}

@misc{podell2023sdxl,
      title={SDXL: Improving Latent Diffusion Models for High-Resolution Image Synthesis}, 
      author={Dustin Podell and Zion English and Kyle Lacey and Andreas Blattmann and Tim Dockhorn and Jonas Müller and Joe Penna and Robin Rombach},
      year={2023},
}

@misc{song2022ddim,
      title={Denoising Diffusion Implicit Models}, 
      author={Jiaming Song and Chenlin Meng and Stefano Ermon},
      year={2022},
}

@inproceedings{lee2023trajcurve,
  title={Minimizing trajectory curvature of ode-based generative models},
  author={Lee, Sangyun and Kim, Beomsu and Ye, Jong Chul},
  booktitle={International Conference on Machine Learning},
  pages={18957--18973},
  year={2023},
  organization={PMLR}
}

@article{kim2024sfno,
  title={Simulation-free training of neural odes on paired data},
  author={Kim, Semin and Yoo, Jaehoon and Kim, Jinwoo and Cha, Yeonwoo and Kim, Saehoon and Hong, Seunghoon},
  journal={Advances in Neural Information Processing Systems},
  volume={37},
  pages={60212--60236},
  year={2024}
}

@misc{ma2025sop,
      title={Inference-Time Scaling for Diffusion Models beyond Scaling Denoising Steps}, 
      author={Nanye Ma and Shangyuan Tong and Haolin Jia and Hexiang Hu and Yu-Chuan Su and Mingda Zhang and Xuan Yang and Yandong Li and Tommi Jaakkola and Xuhui Jia and Saining Xie},
      year={2025},
      eprint={2501.09732},
      archivePrefix={arXiv},
      primaryClass={cs.CV},
}

@misc{luo2026lookahead,
      title={Look-Ahead and Look-Back Flows: Training-Free Image Generation with Trajectory Smoothing}, 
      author={Yan Luo and Henry Huang and Todd Y. Zhou and Mengyu Wang},
      year={2026},
}

@article{krizhevsky2009cifar10,
  title={Learning multiple layers of features from tiny images},
  author={Krizhevsky, Alex and Hinton, Geoffrey and others},
  year={2009},
  publisher={Toronto, ON, Canada}
}

@inproceedings{deng2009imagenet,
  author       = {Jia Deng and
                  Wei Dong and
                  Richard Socher and
                  Li{-}Jia Li and
                  Kai Li and
                  Li Fei{-}Fei},
  title        = {ImageNet: {A} large-scale hierarchical image database},
  booktitle    = {2009 {IEEE} Computer Society Conference on Computer Vision and Pattern
                  Recognition {(CVPR} 2009), 20-25 June 2009, Miami, Florida, {USA}},
  pages        = {248--255},
  publisher    = {{IEEE} Computer Society},
  year         = {2009},
}

@misc{xu2023imagereward,
      title={ImageReward: Learning and Evaluating Human Preferences for Text-to-Image Generation}, 
      author={Jiazheng Xu and Xiao Liu and Yuchen Wu and Yuxuan Tong and Qinkai Li and Ming Ding and Jie Tang and Yuxiao Dong},
      year={2023},
}

@misc{wu2023hpsv2,
      title={Human Preference Score v2: A Solid Benchmark for Evaluating Human Preferences of Text-to-Image Synthesis}, 
      author={Xiaoshi Wu and Yiming Hao and Keqiang Sun and Yixiong Chen and Feng Zhu and Rui Zhao and Hongsheng Li},
      year={2023},
}

@misc{schuhmann2022aesthetic,
      title={LAION-5B: An open large-scale dataset for training next generation image-text models}, 
      author={Christoph Schuhmann and Romain Beaumont and Richard Vencu and Cade Gordon and Ross Wightman and Mehdi Cherti and Theo Coombes and Aarush Katta and Clayton Mullis and Mitchell Wortsman and Patrick Schramowski and Srivatsa Kundurthy and Katherine Crowson and Ludwig Schmidt and Robert Kaczmarczyk and Jenia Jitsev},
      year={2022}
}

@misc{hessel2022clipscore,
      title={CLIPScore: A Reference-free Evaluation Metric for Image Captioning}, 
      author={Jack Hessel and Ari Holtzman and Maxwell Forbes and Ronan Le Bras and Yejin Choi},
      year={2022},
}

@misc{ho2022cfg,
      title={Classifier-Free Diffusion Guidance}, 
      author={Jonathan Ho and Tim Salimans},
      year={2022},
}

@article{lu2022dpm,
  title={Dpm-solver: A fast ode solver for diffusion probabilistic model sampling in around 10 steps},
  author={Lu, Cheng and Zhou, Yuhao and Bao, Fan and Chen, Jianfei and Li, Chongxuan and Zhu, Jun},
  journal={Advances in neural information processing systems},
  volume={35},
  pages={5775--5787},
  year={2022}
}

@article{liu2024audioldm2,
  title={Audioldm 2: Learning holistic audio generation with self-supervised pretraining},
  author={Liu, Haohe and Yuan, Yi and Liu, Xubo and Mei, Xinhao and Kong, Qiuqiang and Tian, Qiao and Wang, Yuping and Wang, Wenwu and Wang, Yuxuan and Plumbley, Mark D},
  journal={IEEE/ACM Transactions on Audio, Speech, and Language Processing},
  volume={32},
  pages={2871--2883},
  year={2024},
  publisher={IEEE}
}

@inproceedings{rombach2022ldm,
  title={High-resolution image synthesis with latent diffusion models},
  author={Rombach, Robin and Blattmann, Andreas and Lorenz, Dominik and Esser, Patrick and Ommer, Bj{\"o}rn},
  booktitle={Proceedings of the IEEE/CVF conference on computer vision and pattern recognition},
  pages={10684--10695},
  year={2022}
}

@article{singer2022makeavideo,
  title={Make-a-video: Text-to-video generation without text-video data},
  author={Singer, Uriel and Polyak, Adam and Hayes, Thomas and Yin, Xi and An, Jie and Zhang, Songyang and Hu, Qiyuan and Yang, Harry and Ashual, Oron and Gafni, Oran and others},
  journal={arXiv preprint arXiv:2209.14792},
  year={2022}
}

@inproceedings{huang2023makeanaudio,
  title={Make-an-audio: Text-to-audio generation with prompt-enhanced diffusion models},
  author={Huang, Rongjie and Huang, Jiawei and Yang, Dongchao and Ren, Yi and Liu, Luping and Li, Mingze and Ye, Zhenhui and Liu, Jinglin and Yin, Xiang and Zhao, Zhou},
  booktitle={International Conference on Machine Learning},
  pages={13916--13932},
  year={2023},
  organization={PMLR}
}

@article{ho2022imagenvideo,
  title={Imagen video: High definition video generation with diffusion models},
  author={Ho, Jonathan and Chan, William and Saharia, Chitwan and Whang, Jay and Gao, Ruiqi and Gritsenko, Alexey and Kingma, Diederik P and Poole, Ben and Norouzi, Mohammad and Fleet, David J and others},
  journal={arXiv preprint arXiv:2210.02303},
  year={2022}
}

@article{xie2024sana,
  title={Sana: Efficient high-resolution image synthesis with linear diffusion transformers},
  author={Xie, Enze and Chen, Junsong and Chen, Junyu and Cai, Han and Tang, Haotian and Lin, Yujun and Zhang, Zhekai and Li, Muyang and Zhu, Ligeng and Lu, Yao and others},
  journal={arXiv preprint arXiv:2410.10629},
  year={2024}
}

@article{heusel2017fid,
  title={Gans trained by a two time-scale update rule converge to a local nash equilibrium},
  author={Heusel, Martin and Ramsauer, Hubert and Unterthiner, Thomas and Nessler, Bernhard and Hochreiter, Sepp},
  journal={Advances in neural information processing systems},
  volume={30},
  year={2017}
}

@article{salimans2016is,
  title={Improved techniques for training gans},
  author={Salimans, Tim and Goodfellow, Ian and Zaremba, Wojciech and Cheung, Vicki and Radford, Alec and Chen, Xi},
  journal={Advances in neural information processing systems},
  volume={29},
  year={2016}
}

@inproceedings{Elson2007cat,
  title={Asirra: a CAPTCHA that exploits interest-aligned manual image categorization},
  author={Jeremy Elson and John R. Douceur and Jon Howell and Jared Saul},
  booktitle={Conference on Computer and Communications Security},
  year={2007},
}

@inproceedings{zhang2018lpips,
  title={The unreasonable effectiveness of deep features as a perceptual metric},
  author={Zhang, Richard and Isola, Phillip and Efros, Alexei A and Shechtman, Eli and Wang, Oliver},
  booktitle={Proceedings of the IEEE conference on computer vision and pattern recognition},
  pages={586--595},
  year={2018}
}

@article{bai2024zigzag,
  title={Zigzag diffusion sampling: Diffusion models can self-improve via self-reflection},
  author={Bai, Lichen and Shao, Shitong and Zhou, Zikai and Qi, Zipeng and Xu, Zhiqiang and Xiong, Haoyi and Xie, Zeke},
  journal={arXiv preprint arXiv:2412.10891},
  year={2024}
}

@inproceedings{wang2025goldencfg,
    title={Towards a Golden Classifier-Free Guidance Path via Foresight Fixed Point Iterations},
    author={Kaibo Wang and Jianda Mao and Tong Wu and Yang Xiang},
    booktitle={The Thirty-ninth Annual Conference on Neural Information Processing Systems},
    year={2025},
}

@inproceedings{
    he2024manifold,
    title={Manifold Preserving Guided Diffusion},
    author={Yutong He and Naoki Murata and Chieh-Hsin Lai and Yuhta Takida and Toshimitsu Uesaka and Dongjun Kim and Wei-Hsiang Liao and Yuki Mitsufuji and J Zico Kolter and Ruslan Salakhutdinov and Stefano Ermon},
    booktitle={The Twelfth International Conference on Learning Representations},
    year={2024},
}

@article{hutchinson89hutchinson,
author = {Hutchinson, M.F.},
year = {1989},
month = {01},
pages = {1059-1076},
title = {A stochastic estimator of the trace of the influence matrix for Laplacian smoothing splines},
volume = {18},
journal = {Communication in Statistics- Simulation and Computation},
}

@misc{ma2024sit,
      title={SiT: Exploring Flow and Diffusion-based Generative Models with Scalable Interpolant Transformers}, 
      author={Nanye Ma and Mark Goldstein and Michael S. Albergo and Nicholas M. Boffi and Eric Vanden-Eijnden and Saining Xie},
      year={2024},
}

@inproceedings{lin2014mscoco,
  title={Microsoft coco: Common objects in context},
  author={Lin, Tsung-Yi and Maire, Michael and Belongie, Serge and Hays, James and Perona, Pietro and Ramanan, Deva and Doll{\'a}r, Piotr and Zitnick, C Lawrence},
  booktitle={European conference on computer vision},
  pages={740--755},
  year={2014},
  organization={Springer}
}

@misc{BlackForestLabs2024FLUX1,
  author = {{Black Forest Labs}},
  title = {{Announcing Black Forest Labs}},
  howpublished = {{Blog post on Black Forest Labs website}},
  month = aug,
  year = 2024,
  url = {https://blackforestlabs.ai/announcing-black-forest-labs/},
  note = {Accessed: 2025-05-14}
}

@article{fan2025cfgzero,
  title   = {{CFG-Zero*}: Improved Classifier-Free Guidance for Flow Matching Models},
  author  = {Fan, Weichen and others},
  journal = {arXiv preprint},
  year    = {2025}
}
\clearpage
\appendix
\renewcommand{\theHsection}{appendix.\Alph{section}}

\section{Proof of Theorem 1}\label{app:proof}
In this section, we prove Theorem~1 of the main paper.
We begin by recalling the flow matching setup and assumptions in Sec.~\ref{subsec:setup}.
We then establish the auxiliary results needed for the proof in Sec.~\ref{sec:lemmas}.
Finally, we present the proof of Theorem~1 in Sec.~\ref{sec:main_proof}.

\subsection{Setup (Definitions and assumptions)}\label{subsec:setup}
Following the setup in Sec.~\ref{sec:preliminaries}, let $x_0 \sim p_0 := \mathcal{N}(0,I)$ and let $x_1 \sim p_1$.
Let $\alpha,\beta:[0,1]\to\mathbb{R}$ be differentiable schedules, and define the interpolant
\begin{equation}\label{eq:xt_def}
x_t := \alpha_t x_1 + \beta_t x_0,
\end{equation}
where $\alpha_t$ is monotonically increasing in $t$, $\beta_t$ is monotonically decreasing in $t$, and $(\alpha_0,\beta_0)=(0,1), \space (\alpha_1,\beta_1)=(1,0)$,
so that $x_t$ evolves from noise to data.

\subsubsection{Sample-wise velocity $v_t$.}
Differentiating \cref{eq:xt_def} with respect to $t$ (holding $(x_0,x_1)$ fixed) gives
\begin{equation}\label{eq:v_def}
v_t := \frac{\dd x_t}{\dd t} = \dot\alpha_t\,x_1 + \dot\beta_t\,x_0.
\end{equation}
Consequently, the interpolant induces the following conditional probability path:
\begin{equation}\label{eq:xt_given_x1}
p_t(x_t\mid x_1)= \mathcal{N}\big(\alpha_t x_1,\ \beta_t^2 I\big).
\end{equation}

\subsubsection{Optimal marginal velocity under MSE.}
Under the mean squared error (MSE) objective, the optimal marginal velocity field is given by
\begin{equation}\label{eq:u_star_def}
u_t(x_t) \;=\; \mathbb{E}\big[v_t \mid x_t\big].
\end{equation}

\subsection{Gaussian identities for the interpolant}
\label{sec:lemmas}

\begin{lemma}[Conditional probability path score $\nabla_{x_t}\log p_t(x_t \mid x_1)$]\label{lem:cond_score}
Fix $t\in(0,1)$. Then the conditional probability path is given by
\begin{equation} 
p_t(x_t\mid x_1)= \mathcal{N}\big(\alpha_t x_1,\ \beta_t^2 I \big),
\end{equation}
and its score satisfies
\begin{equation}\label{eq:score_xt_given_x1}
\nabla_{x_t}\log p_t(x_t\mid x_1) \;=\; -\frac{x_t-\alpha_t x_1}{\beta_t^2}.
\end{equation}
\end{lemma}

\paragraph{proof.}
From the interpolant
\[
x_t=\alpha_t x_1+\beta_t x_0
\]
and the assumption $x_0\sim\mathcal{N}(0,I)$, it follows that, conditioning on $x_1$, $x_t$ is Gaussian with mean $\alpha_t x_1$ and covariance $\beta_t^2 I$. This proves \cref{eq:xt_given_x1}.\\
Next, by \cref{eq:xt_given_x1}, the log-density takes the form
\[
\log p_t(x_t\mid x_1)
=
-\frac{1}{2\beta_t^2}\|x_t-\alpha_t x_1\|^2 + C,
\]
where $C$ is constant with respect to $x_t$. Differentiating with respect to $x_t$ yields
\[
\nabla_{x_t}\log p_t(x_t\mid x_1)
=
-\frac{x_t-\alpha_t x_1}{\beta_t^2},
\]
which proves \cref{eq:score_xt_given_x1}.

\begin{lemma}[Marginal score $\nabla_{x_t}\log p_t(x_t)$ identity]\label{lem:marginal_score}
Fix $t\in(0,1)$ and assume $\beta_t\neq 0$. Let $p_t$ denote the marginal density of $x_t$:
\[
p_t(x_t)=\int p_t(x_t\mid x_1)\,p_1(x_1)\,d x_1.
\]
Under mild regularity conditions allowing differentiation and integration to be interchanged,
\begin{equation}\label{eq:fisher_identity}
\nabla_{x_t}\log p_t(x_t)
=
\mathbb{E}\!\left[\nabla_{x_t}\log p_t(x_t\mid x_1)\ \middle|\ x_t\right].
\end{equation}
Consequently,
\begin{equation}\label{eq:marginal_score}
\nabla_{x_t}\log p_t(x_t)
=
-\frac{x_t-\alpha_t\,\mathbb{E}[x_1\mid x_t]}{\beta_t^2}.
\end{equation}
\end{lemma}

\paragraph{proof.}
Differentiating
\[
p_t(x_t)=\int p_t(x_t\mid x_1)\,p_1(x_1)\,dx_1
\]
with respect to $x_t$ under the integral sign gives
\[
\nabla_{x_t} p_t(x_t)
=
\int \nabla_{x_t} p_t(x_t\mid x_1)\,p_1(x_1)\,dx_1.
\]
Dividing both sides by $p_t(x_t)$ yields \cref{eq:fisher_identity}. Substituting \cref{eq:score_xt_given_x1} from Lemma~\ref{lem:cond_score} into \cref{eq:fisher_identity} then gives
\cref{eq:marginal_score}.

\begin{lemma}[Posterior score $\nabla_{x_t}\log p_t(x_1\mid x_t)$ identity]\label{lem:posterior_score}
Fix $t\in(0,1)$ and assume $\beta_t\neq 0$. By Bayes' rule,
\begin{equation}\label{eq:posterior_score_bayes}
\nabla_{x_t}\log p_t(x_1\mid x_t)
=
\nabla_{x_t}\log p_t(x_t\mid x_1) - \nabla_{x_t}\log p_t(x_t).
\end{equation}
Combining Lemma~\ref{lem:cond_score} and Lemma~\ref{lem:marginal_score} yields
\begin{equation}\label{eq:posterior_score_centered_x1}
\nabla_{x_t}\log p_t(x_1\mid x_t)
=
\frac{\alpha_t}{\beta_t^2}\Big(x_1-\mathbb{E}[x_1\mid x_t]\Big).
\end{equation}
\end{lemma}

\paragraph{proof.}
Since $p_1(x_1)$ does not depend on $x_t$, Bayes' rule gives \cref{eq:posterior_score_bayes}.
Substitute the conditional score \cref{eq:score_xt_given_x1} from Lemma~\ref{lem:cond_score}
and the marginal score \cref{eq:marginal_score} from Lemma~\ref{lem:marginal_score} into \cref{eq:posterior_score_bayes}
to obtain \cref{eq:posterior_score_centered_x1}.

\subsection{Main Proof} \label{sec:main_proof}
\setcounter{theorem}{0}
\begin{theorem}
For any $t$ such that $\alpha_t\neq 0$, the optimal CFM residual satisfies
\begin{equation}\label{eq:thm1_statement}
\mathcal{L}^*_{\text{CFM}}(x_t, t)
= \mathbb{E} \left[ \left\| u_t(x_t)  - v_t\right\|^2 \;\middle|\; x_t \right]
=  \frac{\dot{\alpha}_t\beta_t - \alpha_t \dot{\beta}_t}{\alpha_t}\Big( \beta_t\nabla_{x_t}\cdot u_t(x_t) - \dot{\beta}_t d \Big),
\end{equation}
where $d$ is the dimensionality of the data.
\end{theorem}

\paragraph{proof.}
Fix $t\in(0,1)$ with $\alpha_t\neq 0$, $\beta_t\neq 0$ (as in Lemmas
\ref{lem:cond_score}--\ref{lem:posterior_score}) and define
\begin{equation}
b_t:=\dot\alpha_t-\alpha_t\frac{\dot\beta_t}{\beta_t}.
\label{eq:def_bt}
\end{equation}
We also note that $b_t \neq 0$ since $\alpha_t$ and $\beta_t$ are strictly monotonic and nonzero.

We first express the posterior score $\nabla_{x_t}\log p_t(x_1\mid x_t)$ using $v_t-u_t$.
From \cref{eq:xt_def} with $\beta_t\neq 0$,
\[
x_0=\frac{x_t-\alpha_t x_1}{\beta_t}.
\]
Substituting into \cref{eq:v_def} gives the affine form
\begin{equation}\label{eq:v_affine}
v_t
=
\underbrace{\frac{\dot\beta_t}{\beta_t}}_{=:a_t}\,x_t
+
\underbrace{\Big(\dot\alpha_t-\alpha_t\frac{\dot\beta_t}{\beta_t}\Big)}_{=:b_t}\,x_1.
\end{equation}
Taking conditional expectation given $x_t$ and using \cref{eq:u_star_def},
\begin{equation}\label{eq:u_affine}
u_t(x_t)=\mathbb{E}[v_t\mid x_t]=a_t x_t + b_t\,\mathbb{E}[x_1\mid x_t].
\end{equation}
Subtracting \cref{eq:u_affine} from \cref{eq:v_affine} yields
\begin{equation}\label{eq:residual_relation}
v_t-u_t(x_t)=b_t\Big(x_1-\mathbb{E}[x_1\mid x_t]\Big).
\end{equation}
Since $b_t\neq 0$,
\begin{equation}\label{eq:centered_x1_via_residual}
x_1-\mathbb{E}[x_1\mid x_t]=\frac{1}{b_t}\big(v_t-u_t(x_t)\big).
\end{equation}
On the other hand, by Lemma~\ref{lem:cond_score} and Lemma~\ref{lem:marginal_score}, Bayes' rule yields
Lemma~\ref{lem:posterior_score}; in particular, \cref{eq:posterior_score_centered_x1} holds. Combining
\cref{eq:centered_x1_via_residual} with \cref{eq:posterior_score_centered_x1} gives
\begin{equation}\label{eq:posterior_score_via_residual}
\nabla_{x_t}\log p_t(x_1\mid x_t)
=
\frac{\alpha_t}{\beta_t^2\,b_t}\big(v_t-u_t(x_t)\big).
\end{equation}
We differentiate 
$u_t(x_t)=\int v_t\,p_t(x_1\mid x_t)\,d x_1$, 
we obtain
\begin{align}
J_{x_t}u_t(x_t)
&=\int (J_{x_t}v_t)\,p_t(x_1\mid x_t)\,d x_1
+\int v_t\,\nabla_{x_t}p_t(x_1\mid x_t)^\top d x_1 \notag\\
&=\mathbb{E}[J_{x_t}v_t\mid x_t]
+
\mathbb{E}\!\left[v_t\big(\nabla_{x_t}\log p_t(x_1\mid x_t)\big)^\top \ \middle|\ x_t\right].
\label{eq:Ju_decomp}
\end{align}
From \cref{eq:v_affine}, $v_t=a_t x_t+b_t x_1$, hence
\begin{equation}\label{eq:Jv_const}
J_{x_t}v_t = a_t I=\frac{\dot\beta_t}{\beta_t} I.
\end{equation}
Substituting \cref{eq:Jv_const} and \cref{eq:posterior_score_via_residual} into \cref{eq:Ju_decomp} yields
\begin{equation}\label{eq:Ju_second_moment_matrix}
J_{x_t}u_t(x_t)
=
\frac{\dot\beta_t}{\beta_t} I
+
\frac{\alpha_t}{\beta_t^2 b_t}\,
\mathbb{E}\!\left[v_t\big(v_t-u_t(x_t)\big)^\top\ \middle|\ x_t\right].
\end{equation}
Define the residual $r_t:=v_t-u_t(x_t)$. Since $\mathbb{E}[r_t\mid x_t]=0$ and $v_t=u_t+r_t$, so
\begin{align}
\mathbb{E}\!\left[v_t r_t^\top\mid x_t\right]
&=
\mathbb{E}\!\left[(u_t+r_t)r_t^\top\mid x_t\right]
=
u_t\,\mathbb{E}[r_t^\top\mid x_t]+\mathbb{E}[r_t r_t^\top\mid x_t]\notag\\
&=
\mathbb{E}[r_t r_t^\top\mid x_t].
\label{eq:vr_equals_rr}
\end{align}
Using \cref{eq:vr_equals_rr} in \cref{eq:Ju_second_moment_matrix} gives
\begin{equation}\label{eq:Ju_rr}
J_{x_t}u_t(x_t)
=
\frac{\dot\beta_t}{\beta_t} I
+
\frac{\alpha_t}{\beta_t^2 b_t}\,
\mathbb{E} \!\left[r_t r_t^\top\ \middle|\ x_t\right].
\end{equation}
Rearranging,
\begin{equation}\label{eq:rr_solved}
\mathbb{E}\!\left[r_t r_t^\top\ \middle|\ x_t\right]
=
\frac{\beta_t^2 b_t}{\alpha_t}
\left(
J_{x_t}u_t(x_t)-\frac{\dot\beta_t}{\beta_t} I
\right).
\end{equation}
Since $\|r_t\|^2=\tr(r_t r_t^\top)$,
\begin{equation}\label{eq:mse_trace}
\mathbb{E}\!\left[\|u_t(x_t)-v_t\|^2\ \middle|\ x_t\right]
=
\operatorname{tr}\,\mathbb{E}\!\left[r_t r_t^\top\ \middle|\ x_t\right].
\end{equation}
Taking $\tr(\cdot)$ on both sides of \cref{eq:rr_solved} and using
$\tr(I)=d$ and $\tr(J_{x_t}u_t)=\nabla_{x_t}\cdot u_t$, we obtain
\begin{equation}\label{eq:mse_intermediate}
\mathbb{E}\!\left[\|u_t(x_t)-v_t\|^2\ \middle|\ x_t\right]
=
\frac{\beta_t^2 b_t}{\alpha_t}
\left(
\nabla_{x_t}\cdot u_t(x_t)-\frac{\dot\beta_t}{\beta_t}d
\right).
\end{equation}
Finally, substituting \cref{eq:def_bt} into \cref{eq:mse_intermediate} yields
\begin{align}
\mathbb{E}\!\left[\|u_t(x_t)-v_t\|^2\ \middle|\ x_t\right]
&=
\frac{\beta_t(\dot\alpha_t\beta_t-\alpha_t\dot\beta_t)}{\alpha_t}
\left(
\nabla_{x_t}\cdot u_t(x_t)-\frac{\dot\beta_t}{\beta_t}d
\right) \\
&=
\frac{\dot{\alpha}_t\beta_t - \alpha_t \dot{\beta}_t}{\alpha_t}
\Big( \beta_t\nabla_{x_t}\cdot u_t(x_t) - \dot{\beta}_t\, d \Big).
\label{eq:final_mse}
\end{align}
Therefore,
\begin{equation*}
\mathcal{L}^*_{\text{CFM}}(x_t, t)
= \mathbb{E} \left[ \left\| u_t(x_t)  - v_t\right\|^2 \;\middle|\; x_t \right]
=  \frac{\dot{\alpha}_t\beta_t - \alpha_t \dot{\beta}_t}{\alpha_t}\Big( \beta_t\nabla_{x_t}\cdot u_t(x_t) - \dot{\beta}_t d \Big). \qed
\end{equation*}


\section{Algorithm} \label{app:fds_algo}
In this section, we provide the detailed algorithm of FDS framework of the main paper.
The pseudo-code for \textsc{Refine}$(\cdot)$ and for the overall FDS framework is provided below.
Algorithm~\ref{algo:refine} details the refinement procedure of FDS, while Algorithm~\ref{algo:pipeline} presents the overall generation pipeline.
\begin{algorithm}[H]
\caption{\textsc{Refine}$(x_t, t; u_\theta)$}
\label{algo:refine}
\begin{algorithmic}[1]
    \State \textbf{Input:} number of candidates $M$, number of iteration $N$, perturbation scale $\sigma_t$, flow model $u_\theta$
    \State $\tilde{x}_t^{0} \gets x_t$
    \For{$n = 1, 2, \ldots, N$}
        \State $x_{t,0}^{n} \gets \tilde{x}_t^{\,n-1}$
        \For{$m = 1, 2, \ldots, M$}
            \State $\xi_{t,m}^{n} \sim \mathcal{N}(0, I)$
            \State $x_{t,m}^{n} \gets \tilde{x}_t^{\,n-1} + \sigma_t \xi_{t,m}^{n}$
        \EndFor
        \State $m_t \gets \arg\min_{m \in \{0,1,\ldots,M\}} \hat{\delta}_t\!\left(x_{t,m}^{n}; u_\theta\right)$ \Comment{$\hat{\delta}_t$ approximates the divergence.}
        \State $\tilde{x}_t^{n} \gets x_{t,m_t}^{n}$
    \EndFor
    \State \Return $\tilde{x}_t^{N}$
\end{algorithmic}
\end{algorithm}
\vspace{-10pt}
\begin{algorithm}[H]
\caption{\textsc{Overall Sampling Pipeline with FDS}}
\label{algo:pipeline}
\begin{algorithmic}[1]
    \State \textbf{Input:} initial state $x_{t_0}$, time grid $0=t_0<t_1<\cdots<t_T=1$, flow model $u_\theta$
    \For{$k = 0,1,\dots,T-1$}
        \State $\tilde{x}_{t_k} \gets \operatorname{Refine}(x_{t_k}, t_k; u_\theta)$
        \State $x_{t_{k+1}} \gets \operatorname{SolverStep}(\tilde{x}_{t_k}, t_k, t_{k+1}; u_\theta)$
    \EndFor
    \State \Return $x_{t_T}$
\end{algorithmic}
\end{algorithm}


\section{Implementation Details} \label{app:implementation}
\subsection{Toy Experiments} \label{app:toy_detail}
\subsubsection{Model Architecture and Configuration}
For the toy experiments, the flow model is parameterized by a 4-layer Multi-Layer Perceptron (MLP). 
At inference, both configurations utilize an Euler ODE solver with $20$ integration steps. 
Throughout the process, the noise scale parameter is fixed at $\sigma_t = 0.3$ for all timesteps $t$.

\subsubsection{Measurement of Ground Truth Discrepancy}
To measure the Ground Truth (GT) discrepancy shown in Fig.~\ref{fig:toy_c}, we utilize the full training dataset. By drawing a sufficiently large number of target data points ($N = 100,000$), we construct an empirical target distribution. Relying on the standard flow matching formulation, we can empirically define the exact optimal marginal velocity field, $u_t(x_t)$, under the assumption that this empirical distribution represents the true target. 
Consequently, $u_t(x_t)$ is analytically formulated as the conditional expectation marginalized over the $N$ empirical target samples:
$$u_t(x_t) := \frac{\sum_{k=1}^N p_t(x_t|x_1^{(k)}) v_t(x_t\mid x_1^{(k)})}{\sum_{j=1}^N p_t(x_t|x_1^{(j)})}, \quad v_t(x_t \mid x_1^{(k)}) = \frac{x_1^{(k)} - x_t}{1-t} $$
where $x_1^{(k)}$ denotes the $k$-th sample from the empirical target distribution.

\subsection{Main Experiments} \label{app:main_detail}
\begin{table}[!t]
\centering
\caption{\textbf{Throughput across different configurations.} Values indicate the wall-clock generation time, measured in seconds per batch (sec./batch).}
\label{tab:throughput}
\small
\setlength{\tabcolsep}{3.5pt}
\begin{tabular}{lccccc}
\toprule
  \multirow{2}{*}{\textbf{Solver}} &  \multirow{2}{*}{\textbf{NFE}} & \multicolumn{2}{c}{\textbf{CIFAR-10}} & \multicolumn{2}{c}{\textbf{ImageNet} $\mathbf{256\times 256}$} \\
\cmidrule(lr){3-4} \cmidrule(lr){5-6}
 &  & 
EDM Cond. & EDM Uncond. &  JiT-B/16 &JiT-L/16 \\
                   
\midrule
Euler               & 50      &    2.50    &    2.23     &  5.87  &  17.05  \\
Euler$^\dagger$     & 77      &  3.83     &    3.76  &   9.11 &  26.37 \\
\arrayrulecolor[HTML]{C0C0C0}\midrule
\arrayrulecolor{black}
Euler + \textbf{FDS}& 50      &  3.46    &     3.47     & 9.06 &  26.11 \\
\midrule
Heun                & 99      &    4.94   &     4.41      &   11.76  &  33.98 \\
Heun$^\dagger$      & 153      &   7.61   &      7.48    &  18.20 &   52.54 \\
\arrayrulecolor[HTML]{C0C0C0}\midrule
\arrayrulecolor{black}
Heun + \textbf{FDS} & 99      &  6.86   &   6.87  & 18.10 & 52.01 \\
\bottomrule
\end{tabular}
\end{table}

\subsubsection{Implementation details}
Regarding the perturbation scale $\sigma_t$, we employ a cosine scheduling strategy uniformly across all experiments. 
For the maximum perturbation scale, denoted as $\sigma_{\max}$, is adjusted depending on the target dataset. Specifically, we set $\sigma_{\max} = 0.2$ for the CIFAR-10 dataset, whereas for ImageNet $256\times256$, we utilize $\sigma_{\max} = 0.01$.
To estimate the divergence, we employed Hutchinson's trace estimator, approximating the trace with a single noise vector.

\subsubsection{Compute-matched baseline}

To establish the wall-clock compute-matched baselines (denoted by $\dagger$ in Tab.~\ref{tab:main}), we measured the generation throughput under a single NVIDIA RTX A6000 GPU.
To ensure accurate and robust measurements, we recorded the average execution time across 64 independent iterations of the full image generation process, utilizing a fixed batch size of 32. 
The detailed computational costs associated with each configuration are summarized in Tab.~\ref{tab:throughput}.

\subsection{Text-to-Image Generation}

For the text-to-image generation experiments (Tab.~\ref{tab:t2i}), we utilized the default ODE solver provided in SD3-Medium~\cite{esser2024sd3}, specifically the Euler solver. 
The sampling process was conducted using 50 denoising steps. 
Consistent with our ImageNet $256\times256$ configuration, we applied a cosine schedule for the noise parameter, setting the maximum perturbation scale to $\sigma_{\max} = 0.01$

\subsection{Inverse Problem}
For the inverse problem experiments, we evaluated our method on Gaussian deblurring and super-resolution $(\times 4)$ tasks. 
We adhered to the experimental protocol established by TFG~\cite{ye2024tfg} and adopted the hyperparameter configurations from their official implementation. 
Specifically, the sampling process was conducted using 100 denoising steps, with 4 iterations performed per denoising step. 
To evaluate the performance across both tasks, we utilized 256 samples from the Cat dataset~\cite{Elson2007cat}. 
Furthermore, the maximum perturbation scale was task-specifically set to $\sigma_{\max} = 0.1$ for Gaussian deblurring and $\sigma_{\max} = 0.5$ for super-resolution.

\subsection{HRF Capacity-matched Configuration}
To provide further details regarding the HRF~\cite{zhang2025hrm} configurations presented in Tab.~\ref{tab:traning-based}, we outline the training and architectural specifications for the baseline model.
The HRF model was trained for a total of 400,000 steps, targeting unconditional CIFAR-10 generation.
To ensure a fair comparison by matching the total parameter count (56.0M) with our backbone model, EDM (55.7M), we implemented specific architectural modifications. 
Specifically, we augmented the standard architecture by integrating two additional residual blocks into the U-Net and incorporating five extra linear layers into the time conditioning module.

\begin{figure}[!t]
  \centering
\includegraphics[width=0.8\linewidth]{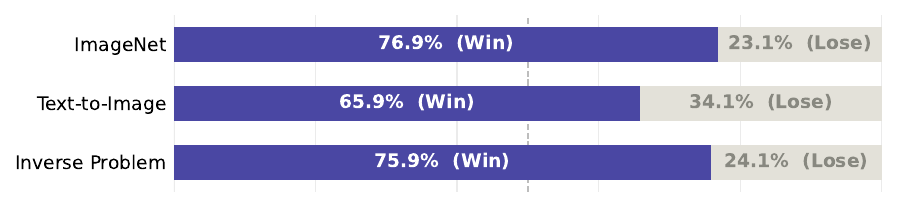}
\captionof{figure}{Comparison of win-rates on user study on ImageNet, text-to-image synthesis and inverse problem.}
\label{fig:win_lose_bar}
\end{figure}


\section{Reliability of Surrogate} \label{app:reliability}
This section evaluates the reliability of employing model divergence as a surrogate for the marginal velocity divergence, as introduced in Sec.~\ref{sec:surrogate}. 
While the synthetic experiment empirically demonstrates that the surrogate accurately captures the ambiguity inherent in real data, we further confirm its effectiveness in practical high-dimensional settings.
To verify the stability of this surrogate across both 2D distributions and high-dimensional image data, we measure the pairwise ordering agreement between the surrogate discrepancy $\hat{\delta}$ and the true discrepancy $\delta^*$.
The agreement probability is formulated as

\begin{equation}
    \Pr\bigl[\operatorname{sign}(\delta^*(x_t) - \delta^*(x_t{+}\sigma_t\xi)) = \operatorname{sign}(\hat{\delta}(x_t) - \hat{\delta}(x_t{+}\sigma_t\xi))\bigr]
\end{equation}

\begin{wraptable}{r}{0.38\columnwidth}
\centering
\vspace{-31pt}
\caption{Agreement of true marginal and surrogate.}
\vspace{2pt}
\label{tab:dis_acc}
\begin{tabular}{lc}
\toprule
\textbf{Dataset} & \textbf{Acc.} (\%) \\
\midrule
2D Synthetic & 89.82 \\
CIFAR-10 & 82.19 \\
\bottomrule
\vspace{-4pt}
\end{tabular}
\end{wraptable}

where $x_t$ and $x_t{+}\sigma_t\xi$ denote the interpolant and perturbed states, respectively.
We compute this agreement across 50K sampled pairs from a synthetic checkerboard distribution and the CIFAR-10 dataset. The results in Tab.~\ref{tab:dis_acc} show an 82\% agreement on CIFAR-10. 
This confirms that our proposed surrogate remains a robust proxy even in high-dimensional settings, demonstrating its suitability for complex image distributions.

While this 18\% error rate indicates potential failure cases, FDS restricts perturbations \textit{locally} to bound trajectory deviations. This constraint prevents severe off-target generation and ensures overall improvements. Our extensive empirical results further support this mechanism, demonstrating that despite the possibility of minor localized errors, the overall generation quality consistently improves across all evaluated configurations.

\section{More Experiments Results} \label{app:more_exps}

\subsection{User Study on Perceptual Quality}

In addition to the quantitative metrics presented in the main manuscript, we conduct a human evaluation to verify the visual improvements provided by FDS. 
Specifically, we designed a comprehensive user study involving 44 participants to assess the tasks discussed in the experiments section. For this study, we randomly sampled image pairs consisting of a baseline generation and its corresponding FDS-applied version. 
The evaluation set included 20 pairs for ImageNet generation, 20 pairs for text-to-image synthesis, and 10 pairs for inverse problems. 
As shown in Fig.~\ref{fig:win_lose_bar}, images generated with FDS achieved an average win-rate of 72.9\%. 
This result demonstrates that the perceptual improvements are significant and consistent across all experimental setups.

\begin{table}[!t]
\centering
\caption{\textbf{Quantitative comparison on CIFAR-10 and ImageNet $\mathbf{256\times256}$.} Results are evaluated using 100 denoising steps.}
\label{tab:100steps}
\scriptsize

\setlength{\tabcolsep}{3.5pt}
\begin{tabular}{lccccccccc}
\toprule
                        &   & \multicolumn{4}{c}{\textbf{CIFAR-10}} & \multicolumn{4}{c}{\textbf{ImageNet} $\mathbf{256\times 256}$} \\
\cmidrule(lr){3-6} \cmidrule(lr){7-10}
\multicolumn{1}{c}{\textbf{Solver}} &\multicolumn{1}{c}{\textbf{NFE}}  & \multicolumn{2}{c}{\textbf{EDM Cond.}} & \multicolumn{2}{c}{\textbf{EDM Uncond.}} &  \multicolumn{2}{c}{\textbf{JiT-B/16}} & \multicolumn{2}{c}{\textbf{JiT-L/16}} \\
\cmidrule(lr){3-4} \cmidrule(lr){5-6}\cmidrule(lr){7-8} \cmidrule(lr){9-10}
                           & & FID ($\downarrow$)  & IS ($\uparrow$)  & FID ($\downarrow$)  & IS ($\uparrow$) &  FID($\downarrow$) & IS ($\uparrow$) & FID ($\downarrow$) & IS ($\uparrow$) \\
\midrule
Euler               & 100      &     2.352  &    9.711     &  2.396  & 9.493    &    3.859      &     277.09   & 2.986    &  336.42 \\
Euler$^\dagger$     & 154      &    2.174  & 9.753  & 2.238 & \textbf{9.543}  & 3.857 &   \textbf{278.75}  &    2.965  & 335.81 \\
\arrayrulecolor[HTML]{C0C0C0}\midrule
\arrayrulecolor{black}
Euler + \textbf{FDS}& 100      & \textbf{1.907}  &   \textbf{9.805}     & \textbf{2.056} & 9.519 & \textbf{3.579} & 278.16 & \textbf{2.741} & \textbf{336.57} \\
\midrule
Heun                & 199      &   1.921    &    9.818    &  2.038  &  9.606 &    3.739    & \textbf{268.29}    &  2.811    &  327.65\\
Heun$^\dagger$      & 307     &    1.924   &  9.816     &   2.042 & 9.603  &   3.833  & 268.12   &    2.875   & \textbf{330.51} \\
\arrayrulecolor[HTML]{C0C0C0}\midrule
\arrayrulecolor{black}
Heun + \textbf{FDS} & 199     &  \textbf{1.785}  &   \textbf{9.868}  & \textbf{1.954} & \textbf{9.633} & \textbf{3.446} &  264.76 & \textbf{2.579}  & 329.07 \\
\bottomrule
\end{tabular}
\end{table}

\subsection{CIFAR-10 and ImageNet Generation}
We provide additional results for our main experiments with 100 denoising steps in Tab.~\ref{tab:100steps}, where FDS consistently outperforms the baselines.

To further validate the robustness of FDS, we evaluate it on ImageNet $256\times256$ using the state-of-the-art latent diffusion backbone SiT~\cite{ma2024sit}. Quantitative results are reported in Tab.~\ref{tab:imagenet_sit} and qualitative samples are shown in Fig.~\ref{fig:sit}. 
\begin{table*}[!t]
\centering
\caption{\textbf{Quantitative comparison on ImageNet $\mathbf{256\times256}$ on SiT-XL.}}
\vspace{-12pt}
\label{tab:imagenet_sit}
\scriptsize

\begin{subtable}[t]{0.48\textwidth}
\centering
\caption{Results with 50 denoising steps.}
\vspace{-8pt}
\label{tab:imagenet_sit_50}
\setlength{\tabcolsep}{4.2pt}
\begin{tabular}{lccc}
\toprule
\textbf{Solver} & \textbf{NFE} & \textbf{FID} ($\downarrow$) & \textbf{IS} ($\uparrow$) \\
\midrule
Euler               & 50  & 2.417 & 249.42 \\
Euler$^\dagger$     & 77  & 2.292 & \textbf{252.15} \\
\arrayrulecolor[HTML]{C0C0C0}\midrule
\arrayrulecolor{black}
Euler + \textbf{FDS}& 50  & \textbf{2.291} & 249.55 \\
\midrule
Heun                & 99  & 2.153 & 255.57 \\
Heun$^\dagger$      & 153 & 2.144 & \textbf{255.58} \\
\arrayrulecolor[HTML]{C0C0C0}\midrule
\arrayrulecolor{black}
Heun + \textbf{FDS} & 99  & \textbf{2.103} & 255.11 \\
\bottomrule
\end{tabular}
\end{subtable}
\hfill
\begin{subtable}[t]{0.48\textwidth}
\centering
\caption{Results with 100 denoising steps.}
\vspace{-8pt}
\label{tab:imagenet_sit_100}
\setlength{\tabcolsep}{4.2pt}
\begin{tabular}{lccc}
\toprule
\textbf{Solver} & \textbf{NFE} & \textbf{FID} ($\downarrow$) & \textbf{IS} ($\uparrow$) \\
\midrule
Euler               & 100 & 2.247 & 253.33 \\
Euler$^\dagger$     & 154 & 2.209 & \textbf{254.69} \\
\arrayrulecolor[HTML]{C0C0C0}\midrule
\arrayrulecolor{black}
Euler + \textbf{FDS}& 100 & \textbf{2.197} & 253.19 \\
\midrule
Heun                & 199 & 2.139 & 255.91 \\
Heun$^\dagger$      & 307 & 2.136 & \textbf{256.18} \\
\arrayrulecolor[HTML]{C0C0C0}\midrule
\arrayrulecolor{black}
Heun + \textbf{FDS} & 199 & \textbf{2.112} & 255.74 \\
\bottomrule
\end{tabular}
\end{subtable}
\end{table*}

Interestingly, the refinement behavior of FDS appears to vary by domain. 
In pixel diffusion models, FDS tends to enrich fine-grained details, whereas in latent diffusion models such as SiT, the corrections appear more structural, possibly reflecting differences in how trajectory crossings manifest across the two spaces.

\subsection{Text-to-Image Synthesis}
To further demonstrate the consistent effectiveness of FDS across diverse text-to-image generation setups, we provide additional experiments scaling our approach to different architectures and evaluation datasets.

\subsubsection{Evaluating on Additional Generative Backbone }
We evaluate the performance of FDS on an additional generative backbone, specifically SD3.5-M~\cite{esser2024sd3}, maintaining the same text-to-image experimental setup. As shown in Tab.~\ref{tab:t2i_backbones}, FDS consistently improves generation quality across most reward metrics. These results confirm that our method is highly generalizable and remains effective across different backbones.

\subsubsection{Evaluating Fidelity on MS-COCO}
Furthermore, we assess the zero-shot generalizability of FDS by evaluating it on the MS-COCO~\cite{lin2014mscoco} dataset to measure generation fidelity rather than relying solely on reward models. We evaluate both generation fidelity and text alignment using zero-shot FID and CLIP scores~\cite{hessel2022clipscore}. As demonstrated in Tab.~\ref{tab:fds_coco}, FDS successfully improves the zero-shot FID over the baseline. Together, these consistent gains substantiate our claim that FDS robustly enhances generation fidelity regardless of the target data distribution or underlying architecture.


\begin{table}[t]
\centering
\footnotesize
\begin{minipage}[t]{0.98\linewidth}
\centering
\captionof{table}{FDS on SD3.5-M.}
\vspace{-4.5pt}
\label{tab:t2i_backbones}
\setlength{\tabcolsep}{5.2pt}

\begin{tabular}{lcccc}
\toprule
\textbf{Method} & \textbf{IR} {($\uparrow$)}& \textbf{HPSv2} {($\uparrow$)}& \textbf{Aes.} {($\uparrow$)}& \textbf{CLIP} {($\uparrow$)}\\
\midrule
SD3.5-M \cite{esser2024sd3} & 81.05 & 27.75 & 5.81 & 28.52 \\
SD3.5-M + \textbf{FDS} & \textbf{83.60} & \textbf{27.85} & \textbf{5.82} & \textbf{28.53} \\
\bottomrule
\end{tabular}

\end{minipage}
\vfill
\vspace{6pt}
\begin{minipage}[t]{0.96\linewidth}
\centering
\vspace{2pt}
\captionof{table}{FDS on MS-COCO}
\vspace{-5pt}
\label{tab:fds_coco}

\setlength{\tabcolsep}{5.3pt}
\begin{tabular}{lcccc}
\toprule
\multirow{2}{*}{\textbf{Method}} &
\multicolumn{2}{c}{\textbf{CFG = 3.0}} &
\multicolumn{2}{c}{\textbf{CFG = 7.0}} \\
\cmidrule(lr){2-3}\cmidrule(lr){4-5}
 & FID {($\downarrow$)}& CLIP {($\uparrow$)} & FID {($\downarrow$)}& CLIP {($\uparrow$)}\\
\midrule
SD3-M  & 16.92 & 25.96& 23.11 & 26.29 \\
SD3-M + \textbf{FDS} & \textbf{16.20} & \textbf{26.02} & \textbf{22.18} & \textbf{26.33}  \\
\bottomrule
\end{tabular}
\end{minipage}

\end{table}

\section{More Qualitative Results} \label{app:more_qual}
\begin{figure}[p]
  \centering

  \begin{subfigure}{0.78\linewidth}
    \centering
    \includegraphics[width=\linewidth]{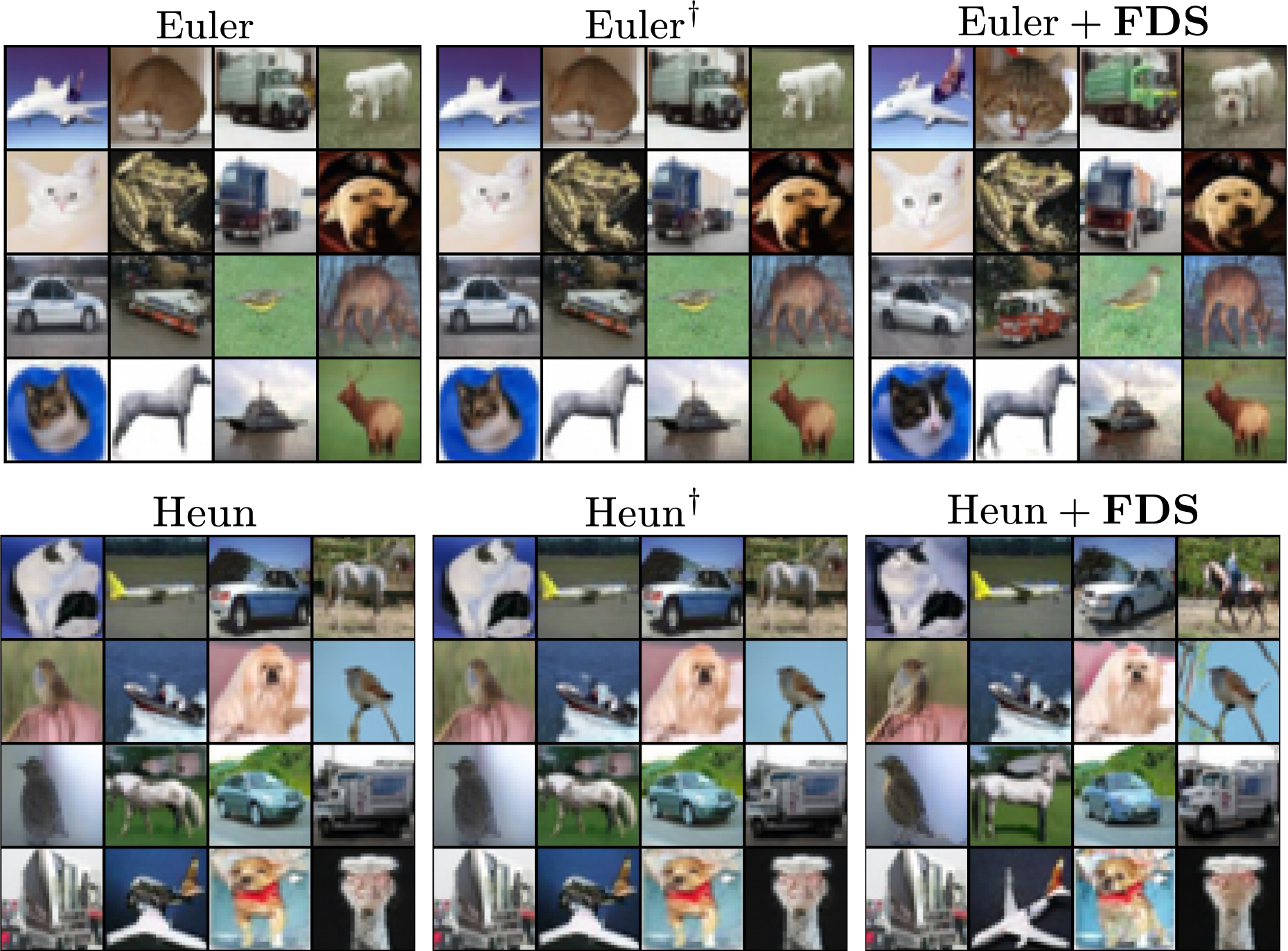}
    \caption{Conditional generation results on CIFAR-10.}
    \label{fig:cifar10_cond}
  \end{subfigure}

  \vspace{0.8cm}

  \begin{subfigure}{0.78\linewidth}
    \centering
    \includegraphics[width=\linewidth]{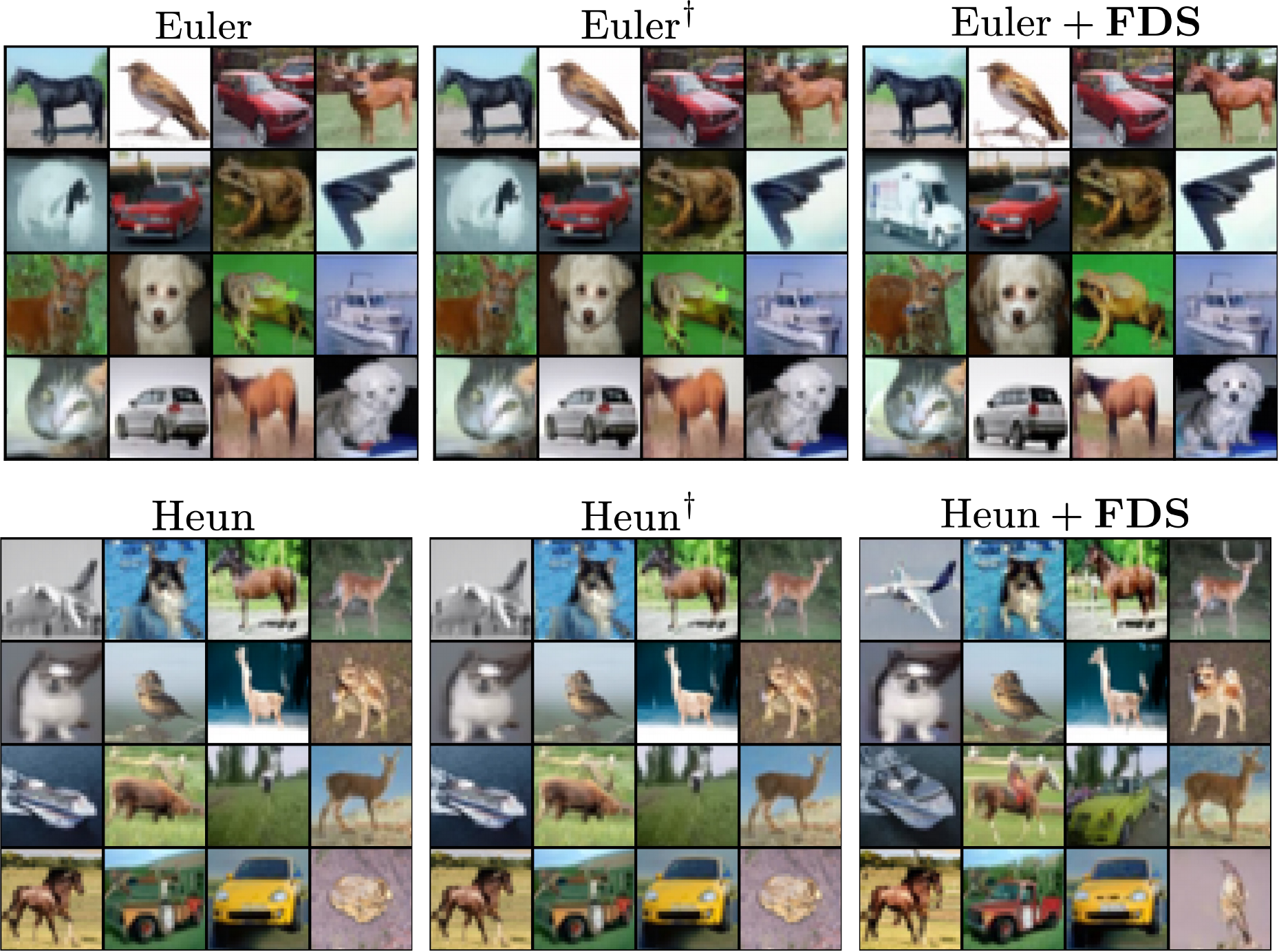}
    \caption{Unconditional generation results on CIFAR-10.}
    \label{fig:cifar10_uncond}
  \end{subfigure}

  \caption{\textbf{Qualitative results on CIFAR-10.} $\dagger$ denotes a base
solver with an increased NFEs to match the wall-clock time of our framework. Compared
to the compute-matched baseline($\dagger$), FDS effectively enhances the generation quality.}
  \label{fig:cifar}
\end{figure}

\begin{figure}[p]
  \centering
  \includegraphics[width=0.99\linewidth]{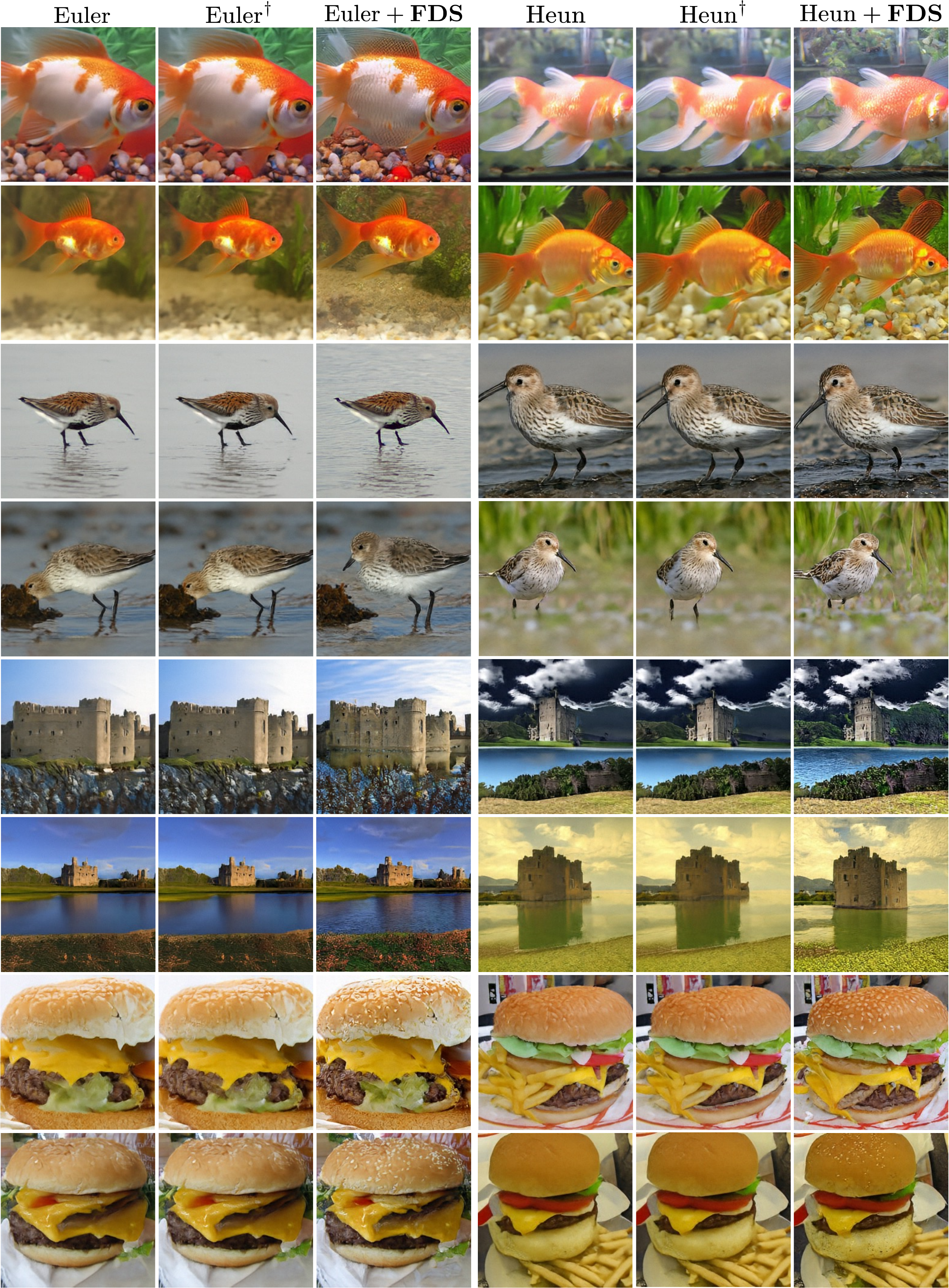}
  \caption{\textbf{Qualitative results on ImageNet $\mathbf{256 \times 256}$.} Generated samples using the JiT-B/16 backbone. Arranged from top to bottom, every two rows show instances from classes 1 (goldfish), 140 (dunlin), 483 (castle), and 933 (cheeseburger).}
  \label{fig:jitb}
\end{figure}

\begin{figure}[p]
  \centering
  \includegraphics[width=0.99\linewidth]{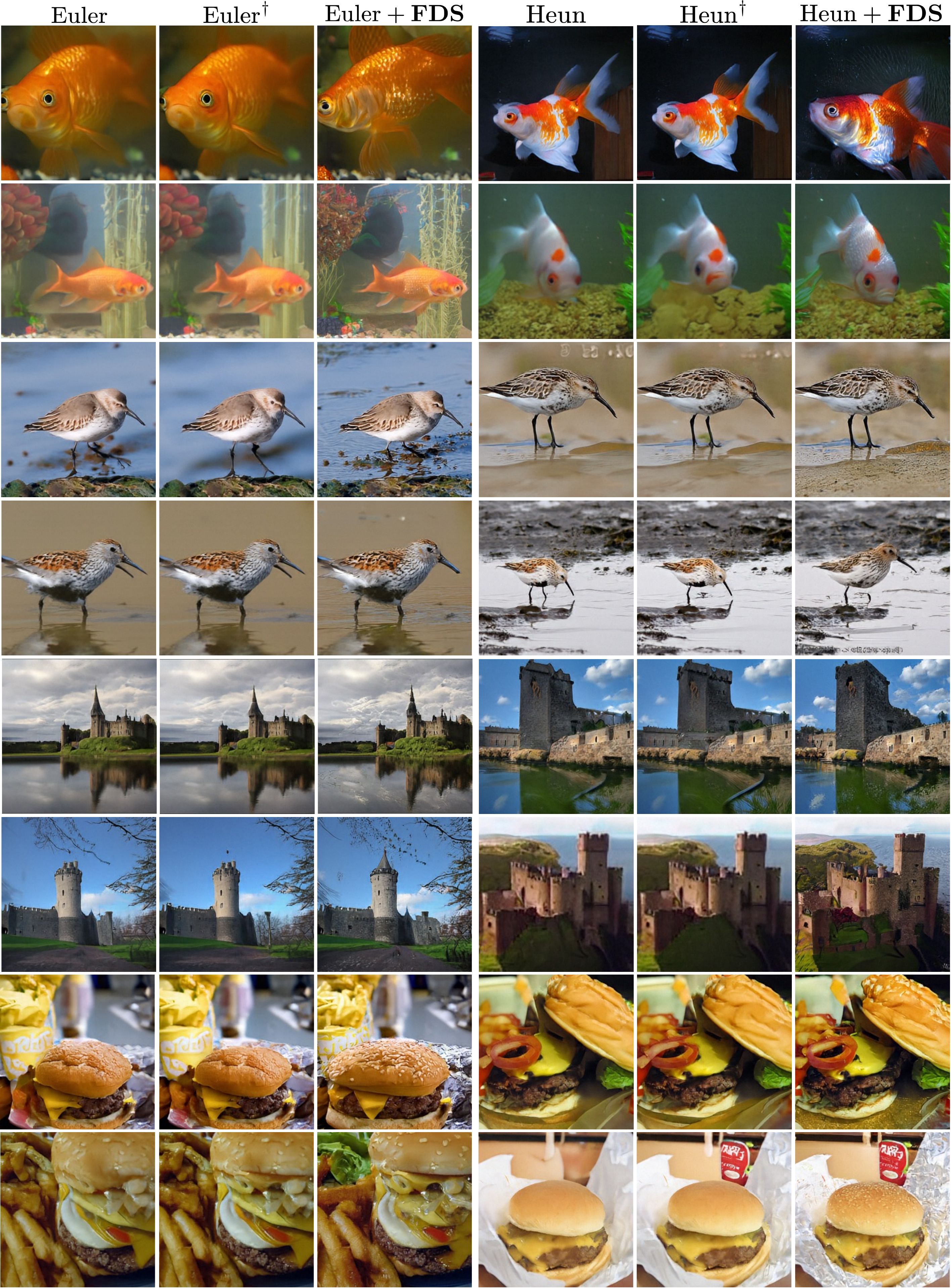}
  \caption{\textbf{Qualitative results on ImageNet $\mathbf{256 \times 256}$.} Generated samples using the JiT-L/16 backbone. Arranged from top to bottom, every two rows show instances from classes 1 (goldfish), 140 (dunlin), 483 (castle), and 933 (cheeseburger).}
  \label{fig:jitl}
\end{figure}

\begin{figure}[p]
  \centering
  \includegraphics[width=0.99\linewidth]{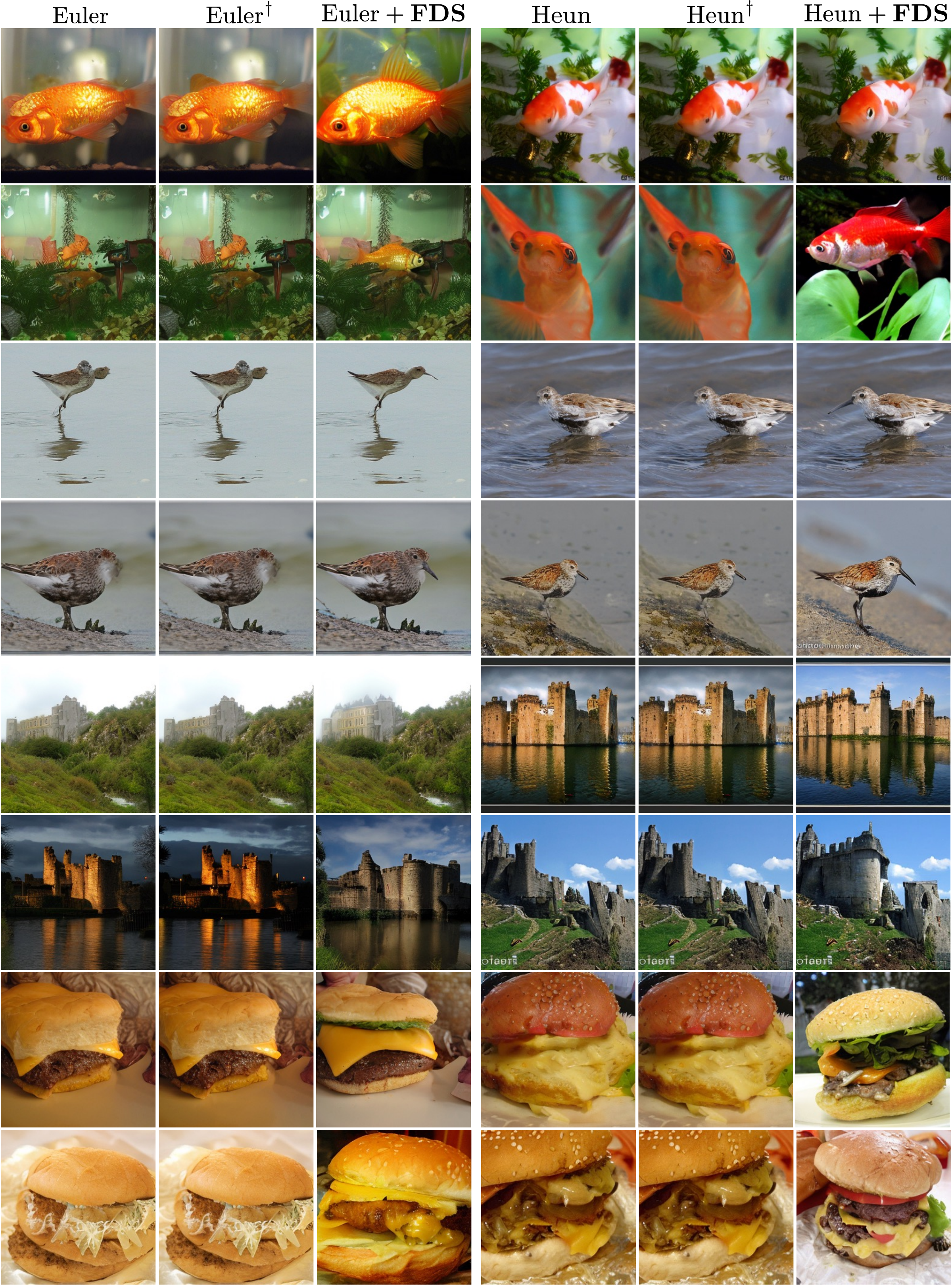}
  \caption{\textbf{Qualitative results on ImageNet $\mathbf{256 \times 256}$.} Generated samples using the SiT-XL backbone. Arranged from top to bottom, every two rows show instances from classes 1 (goldfish), 140 (dunlin), 483 (castle), and 933 (cheeseburger).}
  \label{fig:sit}
\end{figure}

\begin{figure}[p]
  \centering
  \begin{subfigure}[t]{0.46\textwidth}
    \centering
    \includegraphics[width=0.92\linewidth]{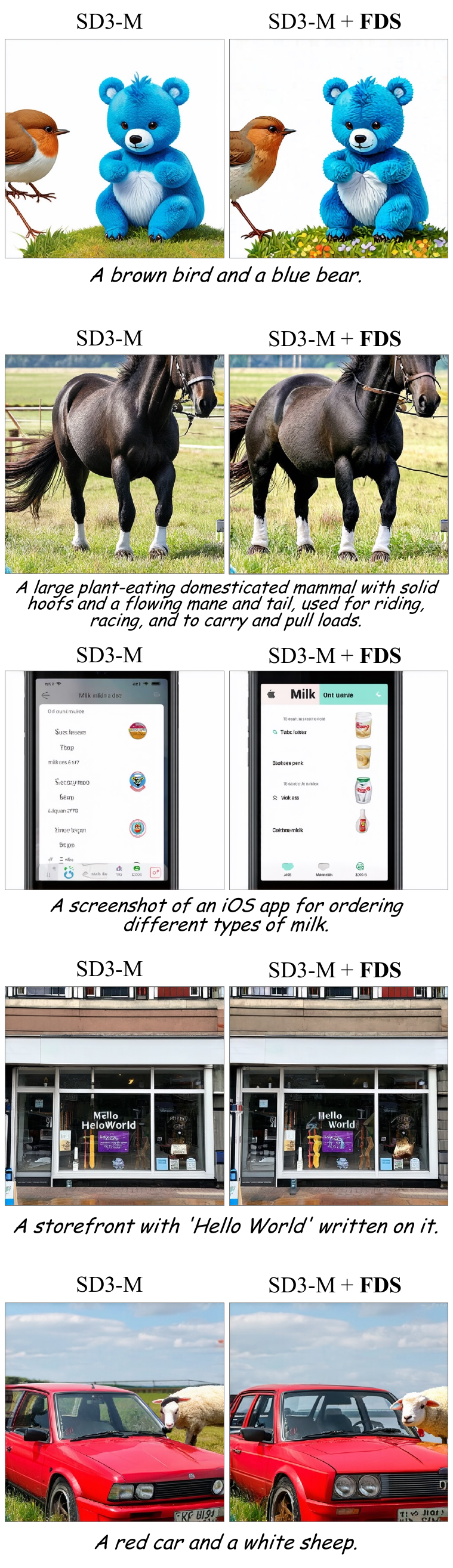}
    \label{fig:t2i_cfg3}
  \end{subfigure}\hfill
  \begin{subfigure}[t]{0.46\textwidth}
    \centering
    \includegraphics[width=0.913\linewidth]{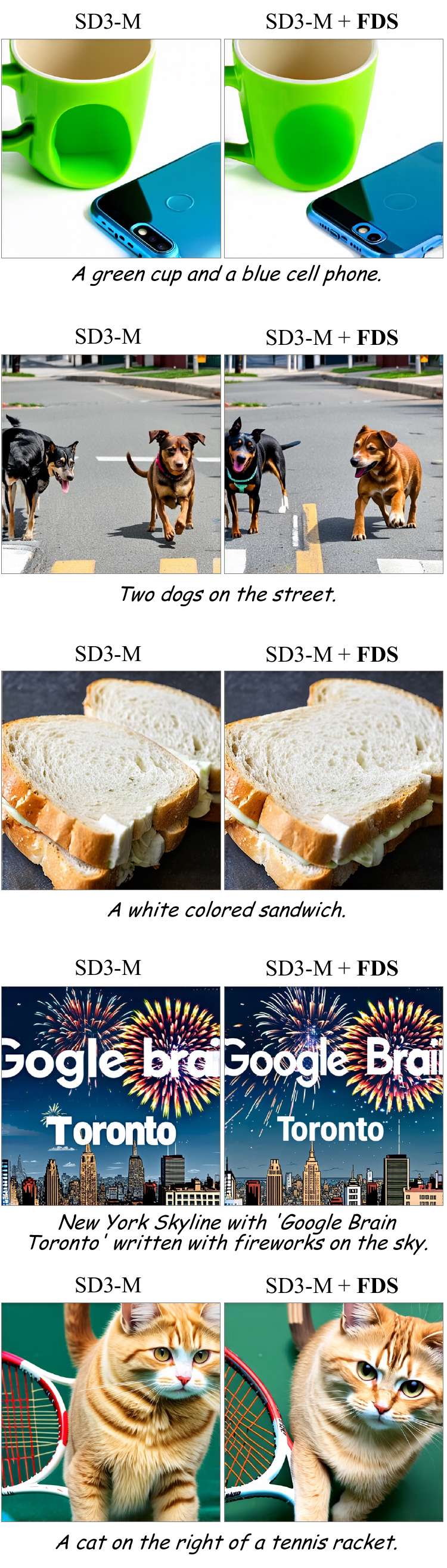}
    \label{fig:t2i_cfg7}
  \end{subfigure}
  \caption{\textbf{Qualitative results for text-to-image generation.}}
  \label{fig:supp_t2i}
\end{figure}

\begin{figure}[p]
  \centering
  \includegraphics[width=0.73\linewidth]{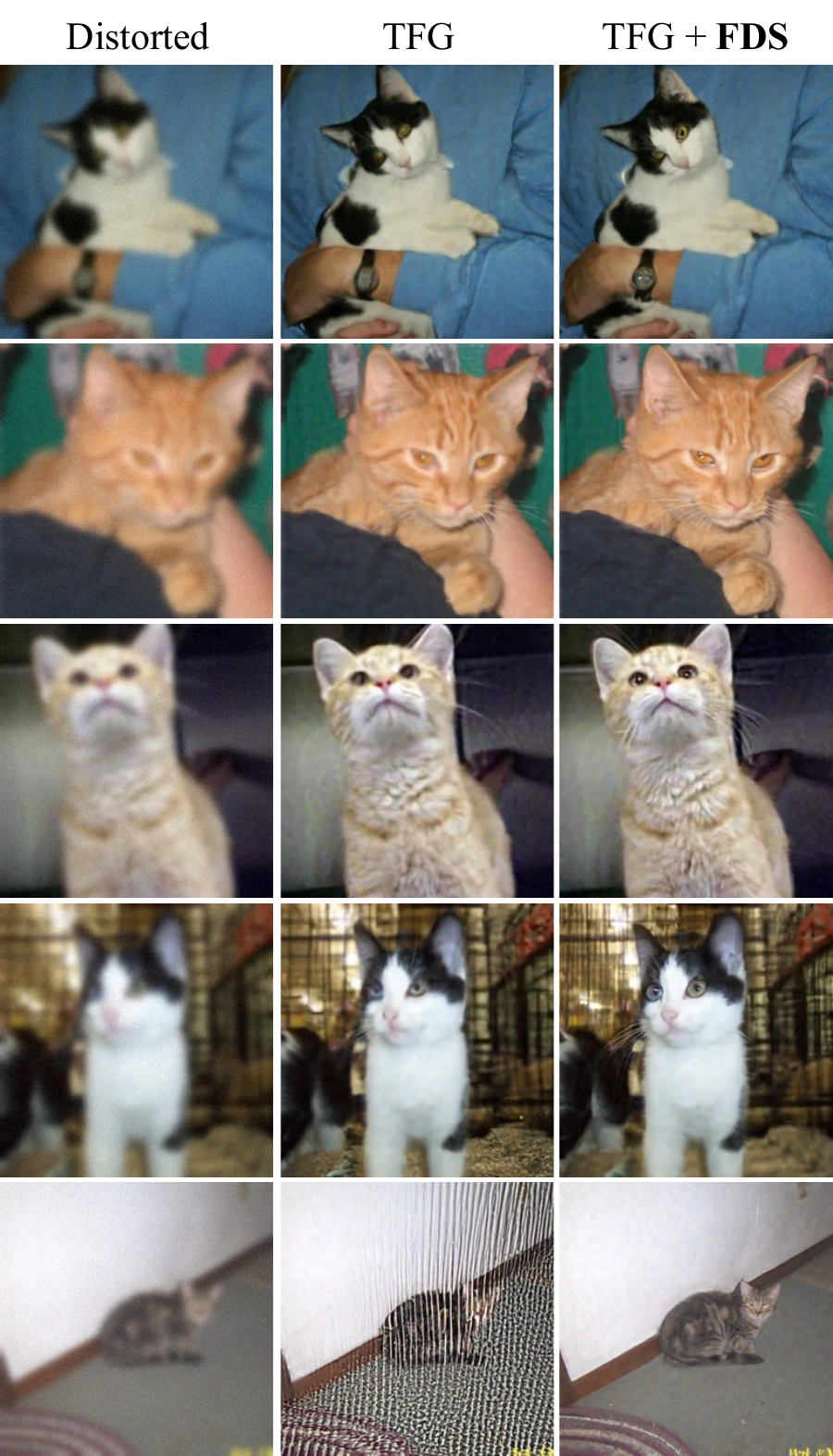}
  \caption{Qualitative results on Gaussian deblurring.}
  \label{fig:supp_deblur}
\end{figure}

\begin{figure}[p]
  \centering
  \includegraphics[width=0.73\linewidth]{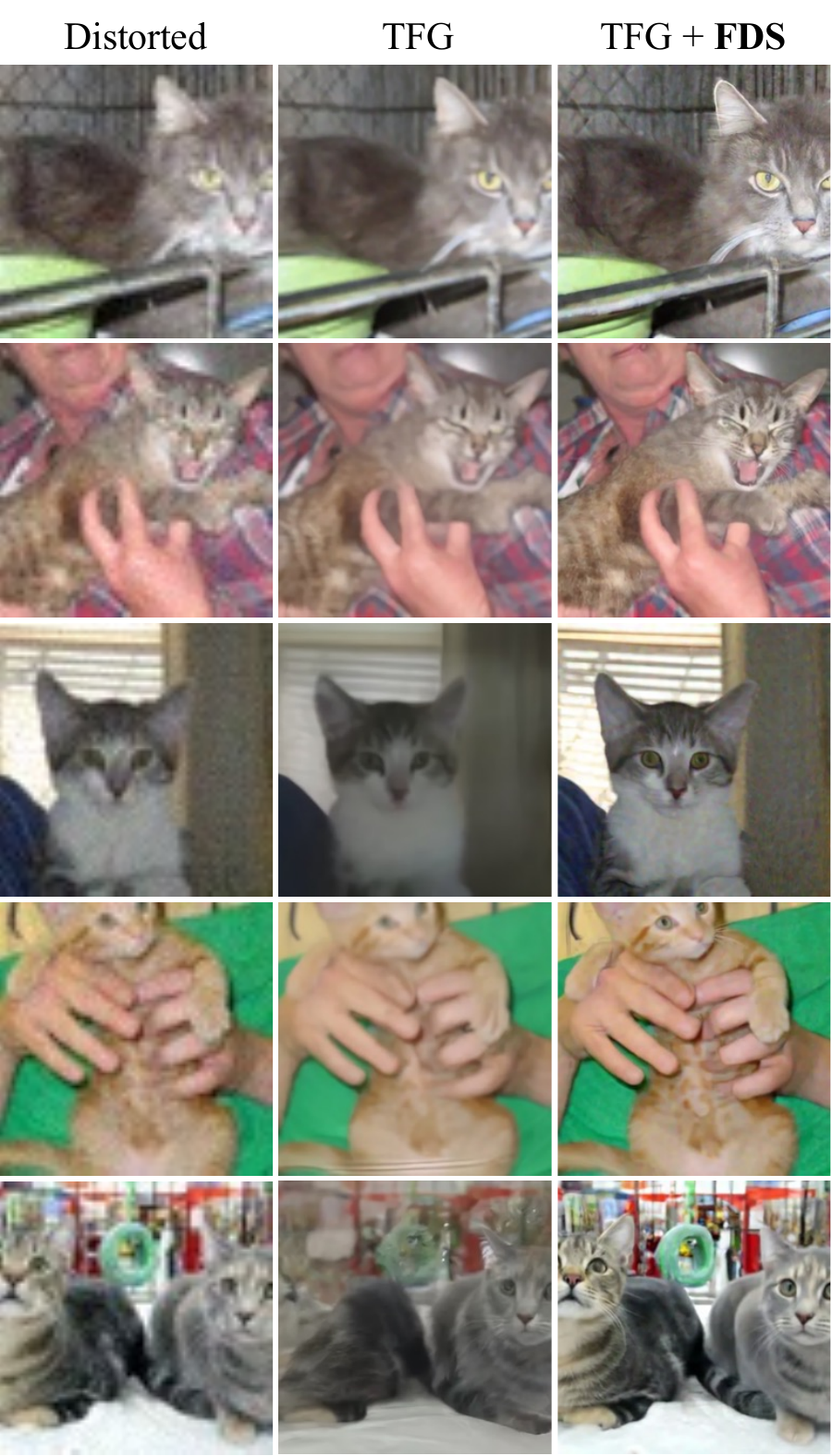}
  \caption{Qualitative results on Super-Resolution $\times 4$}
  \label{fig:supp_sr}
\end{figure}

In this section, we present additional qualitative results on various generation settings, including the CIFAR-10 (Fig.~\ref{fig:cifar}), ImageNet $256\times256$ (Fig.~\ref{fig:jitb}--\ref{fig:sit}), and text-to-image generation (Fig.~\ref{fig:supp_t2i}) and inverse problem (Fig.~\ref{fig:supp_deblur}--\ref{fig:supp_sr}).
As discussed in Sec.~\ref{sec:exp}, FDS consistently improves the generation quality.

\end{document}